\documentclass[11pt,table,dvipsnames]{article}
\usepackage{amsmath}
\usepackage[preprint]{acl}
\usepackage{subcaption}
\usepackage{multirow}
\usepackage{pifont}

\usepackage[most]{tcolorbox}

\usepackage{graphicx}
\usepackage{float}
\usepackage{booktabs}
\usepackage{multirow}
\usepackage{tabularx}
\usepackage{array}

\usepackage{booktabs}
\usepackage{pifont}
\usepackage{array}
\definecolor{DarkGreen}{RGB}{0,100,0}
\usepackage{booktabs}
\usepackage{pifont}
\usepackage{graphicx} % for \resizebox
\usepackage{cuted}
\usepackage{listings}

\tcbuselibrary{breakable}

\tcbuselibrary{breakable,listings}

\newcommand{\cmark}{\textcolor{DarkGreen}{\ding{51}}}
\newcommand{\xmark}{\textcolor{red!70!black}{\ding{55}}}
\newcommand{\pmark}{\textcolor{orange!90!black}{\textbf{\textemdash}}} % Partial

\usepackage{algorithm}
\usepackage{algpseudocode}
\algrenewcommand\algorithmicrequire{\textbf{Input:}}
\algrenewcommand\algorithmicensure{\textbf{Output:}}
\algrenewcommand{\algorithmiccomment}[1]{\hfill$\triangleright$ #1}

\newcounter{myboxcounter}

\newtcolorbox{mybox}[2][]{
    before upper={
        \refstepcounter{myboxcounter}
    },
    colback=cyan!3,
    colframe=cyan!25!blue!75,
    title=\textbf{#2},
    breakable,
    #1
}

\definecolor{darkgreen}{RGB}{0,100,0}   % define the color you're using
\usepackage{booktabs}
\definecolor{A}{RGB}{255,248,220} % light yellow
\definecolor{B}{RGB}{255,235,170} % medium yellow
\definecolor{avggray}{RGB}{235,235,235} 
\definecolor{green}{RGB}{190,225,190} 
\usepackage{mdframed}

\definecolor{rqheader}{RGB}{45,120,75}    % forest green
\definecolor{rqbg}{RGB}{238,247,240}       % light green background

\definecolor{promptheader}{RGB}{40,65,110}      % deep navy
\definecolor{promptbg}{RGB}{240,245,252}        % light blue background
\definecolor{border}{RGB}{120,140,180}    % algorithm border: muted steel-blue

\newmdenv[
  topline=false,
  bottomline=false,
  rightline=false,
  leftline=true,
  linecolor=border,
  linewidth=3pt,
  innertopmargin=4pt,
  innerbottommargin=4pt,
  innerleftmargin=10pt,
  innerrightmargin=4pt,
  skipabove=4pt,
  skipbelow=4pt
]{algobox}

\usepackage{times}
\usepackage{latexsym}

\usepackage[T1]{fontenc}
\usepackage[utf8]{inputenc}

\usepackage{microtype}

\usepackage{inconsolata}

\usepackage{graphicx}

\usepackage[most]{tcolorbox}

\usepackage[most]{tcolorbox}

\definecolor{PreprocessBlue}{HTML}{1565C0}
\definecolor{ChunkOrange}{HTML}{EF6C00}
\definecolor{COPEGreen}{HTML}{2E7D32}
\definecolor{IntentPurple}{HTML}{6A1B9A}
\definecolor{RetrieveRed}{HTML}{C62828}
\definecolor{GenerateTeal}{HTML}{00897B}
\definecolor{VerifyCyan}{HTML}{00838F}

\def\bng{\bngx}

\def\bns{\bnsx}

\font\bngx=bang10

\font\bnsx=bangsl10

\def\*#1*#2{o\null{#2}{#1}}

\def\sh#1{\setbox0=\hbox{#1}%
     \kern-.02em\copy0\kern-\wd0
     \kern.04em\copy0\kern-\wd0
     \kern-.02em\raise.0433em\box0 }
\title{\textbf{\textsc{TeachMateGPT:}} A Multi-Agent Knowledge-Grounded Framework for Pedagogical Assessment Generation from Science Curriculum Materials}

\author{
 \textbf{Fatema Tuj Johora Faria\textsuperscript{1}},
 \textbf{Mukaffi Bin Moin\textsuperscript{1}},
 \textbf{M. F. Mridha\textsuperscript{2}},
 \textbf{Jubayer Al Mahmud\textsuperscript{3}}
\\
\\
 \textsuperscript{1}Ahsanullah University of Science and Technology, Bangladesh\\
 \textsuperscript{2}American International University - Bangladesh\\
 \textsuperscript{3}Jashore University of Science and Technology, Bangladesh\\
 \small{
   \textbf{Correspondence:} \href{mailto:mukaffi28@gmail.com}{mukaffi28@gmail.com}, \href{mailto:fatema.faria142@gmail.com}{fatema.faria142@gmail.com} 
 }
}

\begin{document}
\maketitle

\begin{abstract}

Automatically generating textbook-grounded assessment items can reduce science teachers' workload, but existing retrieval-augmented generation (RAG) systems rely on flat retrieval, support only single-question generation, lack safeguards against weak evidence, and are ill-suited to low-resource, board-exam-structured curricula. We address these limitations with \textbf{\textsc{TeachMateGPT}}, a multi-agent system contributing four advances to curriculum-grounded science-assessment authoring. \textbf{(i)} \textbf{COPE}, a hierarchical knowledge base replacing token-window chunking with a multi-resolution index that segments documents along syllabus structure and links them at three granularities via a traversable graph-based lineage, matching evidence to each topic's instructional level. \textbf{(ii)} A staged, fail-closed agent pipeline replacing one-shot retrieve-then-generate: routing gates search, retrieval fuses dense and lexical evidence under a coverage gate that withholds generation on insufficient evidence, and specialist agents draft objective and constructed-response items. \textbf{(iii)} \textbf{SAVER}, a source-attributed verification protocol scoring faithfulness, relevance, and hallucination risk against retrieved evidence, applying stricter grounding checks across each creative question's four sub-parts, paired with \textbf{teacher-in-the-loop evaluation} rather than automatic filtering. \textbf{(iv)} \textbf{NCTB-SciGen8}, a curriculum-grounded dataset of 198 items (143 multiple-choice, 55 creative questions) spanning all 14 chapters of the NCTB Class 8 science textbook, produced by the pipeline and rated by three practicing teachers. TeachMateGPT raises \textbf{faithfulness} (\textbf{0.68 $\rightarrow$ 0.96}) and \textbf{answer relevancy} (\textbf{0.60 $\rightarrow$ 0.89}) over a vanilla RAG baseline.

\end{abstract}

\section{Introduction} 
\begin{figure}[t]
    \centering
    \includegraphics[width=\linewidth]{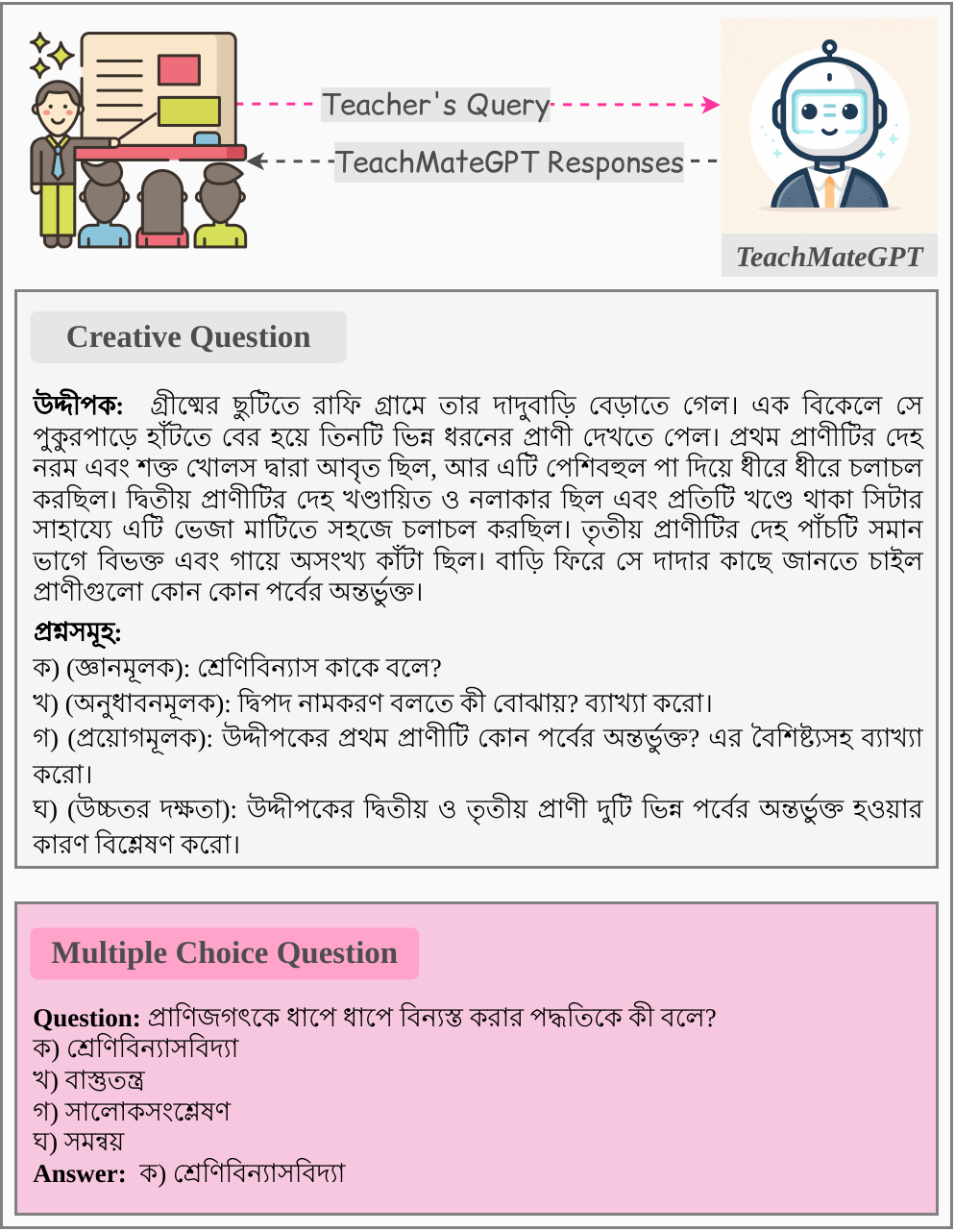}
    \caption{The diagram illustrates the flow from a teacher's natural-language query with a target difficulty level to curriculum-grounded assessment generation. \textbf{\textsc{TeachMateGPT}} retrieves relevant textbook content and generates board-style \textbf{CQs} and \textbf{MCQs} that match the requested topic and difficulty.}
    \label{intro-example}
\end{figure} 

Large language models (LLMs) can accelerate assessment creation, but unconstrained generation may introduce unsupported facts, chapter drift, weak distractors, or invalid board structures. This calls for knowledge-grounded, multi-format generation under pedagogical constraints that retrieves reliable evidence, withholds generation when evidence is insufficient, and provides provenance for teacher verification. This is particularly important for Bangladeshi science teachers who prepare Class~8 assessments from the National Curriculum and Textbook Board (NCTB) \citep{nctb2025} syllabus, where items must align with textbooks, use Bangla, and match the intended difficulty.

Retrieval-augmented generation (RAG) addresses hallucination by coupling an LLM with an external knowledge store, conditioning generation on retrieved passages, and is now common in educational NLP \citep{swacha2025rag,pan2025survey}. Recent RAG-based systems retrieve course materials or exam corpora to generate assessment items and answer keys \citep{pradeesh2025rag,poscomp2025rag,fie2025agentic}, construct mixed-format examinations from domain knowledge bases \citep{jait2025exam}, and support textbook resources such as Bangla B-RAG and NCTB-QA \citep{alawwad2025textbook,brag2025,nctbqa2026}. Other studies improve distractor quality through teacher-student reasoning \citep{biflow2025} and difficulty-controlled generation through knowledge graphs \citep{kaqg2025}. However, existing RAG systems rely on flat retrieval over generic chunks and target general question generation rather than curriculum-specific, board-style assessment authoring. For Bangla NCTB resources, available systems mainly focus on answering questions instead of generating teacher-facing, multi-format assessments with evidence-aware refusal.

While RAG provides grounding, a vanilla retrieve-then-generate pipeline remains insufficient for assessment authoring. Teacher requests may require clarification, retrieval must account for curriculum context and noisy textbook sources, assessment formats impose distinct pedagogical constraints, and generated items require validation before classroom use. Recent educational systems therefore adopt agentic workflows that coordinate routing, retrieval, generation, and verification rather than relying on a single generation step \citep{codegen2026,fie2025agentic,kaqg2025,wang2025ailiteracy,jia2025eduagentqg}. However, existing systems remain English-centric and provide limited support for query clarification, multi-format generation, and source-attributed verification.

Building on these observations, we introduce \textbf{\textsc{TeachMateGPT}}, a multi-agent, curriculum-grounded framework for Bangla Class~8 NCTB science assessment generation, with four contributions. \textbf{Firstly}, \textbf{\textsc{COPE}} (\textbf{C}urriculum-\textbf{O}riented \textbf{P}edagogical \textbf{E}mbedding), a hierarchical curriculum index organizing textbook content at multiple instructional resolutions; unlike generic parent-child chunking, every chunk carries ingest-time lineage and neighbor links that retrieval can traverse to recover related instructional content, and removing COPE produces the largest drop in answer relevancy among the ablated components, along with a substantial reduction in stimulus realism. \textbf{Secondly}, a staged, fail-closed multi-agent pipeline for query routing, hybrid retrieval, evidence refinement, and format-specific generation, incorporating \textbf{\textsc{CCI}} (\textbf{C}ontextual \textbf{C}ontent \textbf{I}njection) for evidence restoration and \textbf{\textsc{CCR}} (\textbf{C}onsensus-Based \textbf{C}onflict \textbf{R}esolution) for redundancy reduction. \textbf{Thirdly}, \textbf{\textsc{SAVER}} (\textbf{S}ource-\textbf{A}ttributed \textbf{V}erification and \textbf{E}vidence \textbf{R}anking), which scores faithfulness, relevance, and hallucination risk before teacher presentation and flags unsupported items for review rather than filtering them automatically. \textbf{Fourthly}, \textbf{NCTB-SciGen8}, a curriculum-grounded evaluation dataset produced directly by the pipeline and reviewed by three practicing science teachers. Together, these four contributions address the following research questions.

\begin{itemize}
    \item \textbf{RQ1.} How can authorized science textbooks be organized into a retrieval-ready curriculum knowledge base that preserves instructional hierarchy for assessment generation?
    \item \textbf{RQ2.} How should teacher queries be safely routed and clarified so that only well-specified curriculum assessment requests proceed to evidence retrieval?
    \item \textbf{RQ3.} How can curriculum evidence be retrieved and refined so that generated assessments remain topic-consistent, contextually complete, and withhold generation when coverage is insufficient? 
    \item \textbf{RQ4.} How can multiple classroom assessment formats, multiple-choice and board-style creative questions, be generated at controllable difficulty while staying grounded in retrieved curriculum evidence? 
    \item \textbf{RQ5.} How can automatic source-attributed verification, combined with a teacher-in-the-loop review process, support trustworthy acceptance, editing, or cautioning of AI-generated assessments before classroom use?
\end{itemize}

{Figure~\ref{intro-example} presents an example interaction with \textbf{\textsc{TeachMateGPT}} and illustrates the multi-agent workflow from a teacher query to curriculum-grounded assessment items.}

\section{Related Work}

\subsection{Retrieval-Augmented Generation for Educational NLP}

RAG connects LLMs with external knowledge sources to incorporate evidence during generation. Recent surveys highlight its growing role in educational applications, particularly for knowledge-intensive tasks that require reliable access to instructional materials \citep{swacha2025rag,pan2025survey}. Existing studies have explored retrieval-based assessment creation from course documents and examination archives, including MCQ generation with answer keys \citep{pradeesh2025rag,poscomp2025rag,fie2025agentic}, mixed-format exam construction from domain knowledge bases \citep{jait2025exam}, and textbook-based educational QA \citep{alawwad2025textbook}. For low-resource educational contexts, Bangla resources such as B-RAG and NCTB-QA provide curriculum-specific retrieval benchmarks and demonstrate the potential of NCTB-grounded educational systems \citep{brag2025,nctbqa2026}. Complementary studies explore reasoning-based strategies for improving distractor quality \citep{biflow2025} and knowledge graph--guided generation with cognitive difficulty control \citep{kaqg2025}. 

\subsection{Agentic Workflows for Educational Assessment Generation}

Recent educational systems increasingly use agent-based architectures to divide complex assessment tasks among specialized components. CODE-GEN presents a human-in-the-loop RAG agent framework for coding-comprehension MCQs, where separate modules handle item creation and quality assessment \citep{codegen2026}. Other approaches distribute educational workflows across agents for document analysis, retrieval, question construction, and evaluation to improve consistency with course content \citep{fie2025agentic}. Knowledge graph enhanced multi-agent RAG frameworks further incorporate cognitive objectives and difficulty calibration through Bloom's taxonomy and Item Response Theory \citep{kaqg2025}. Related studies explore collaborative generation strategies, such as multi-agent MCQ construction and teacher-student reasoning, to improve distractor quality and assessment reliability \citep{tian2026requesta,biflow2025}.

\subsection{Research Gap}

Although recent studies have advanced educational assessment generation, they focus primarily on isolated components of the pipeline. Table~\ref{tab:research_gap} summarizes representative systems and highlights the capabilities missing from existing approaches.

\section{The \textsc{TeachMateGPT} Framework}
\label{sec:framework}

\begin{figure*}[t]
\centering
\includegraphics[width=\textwidth]{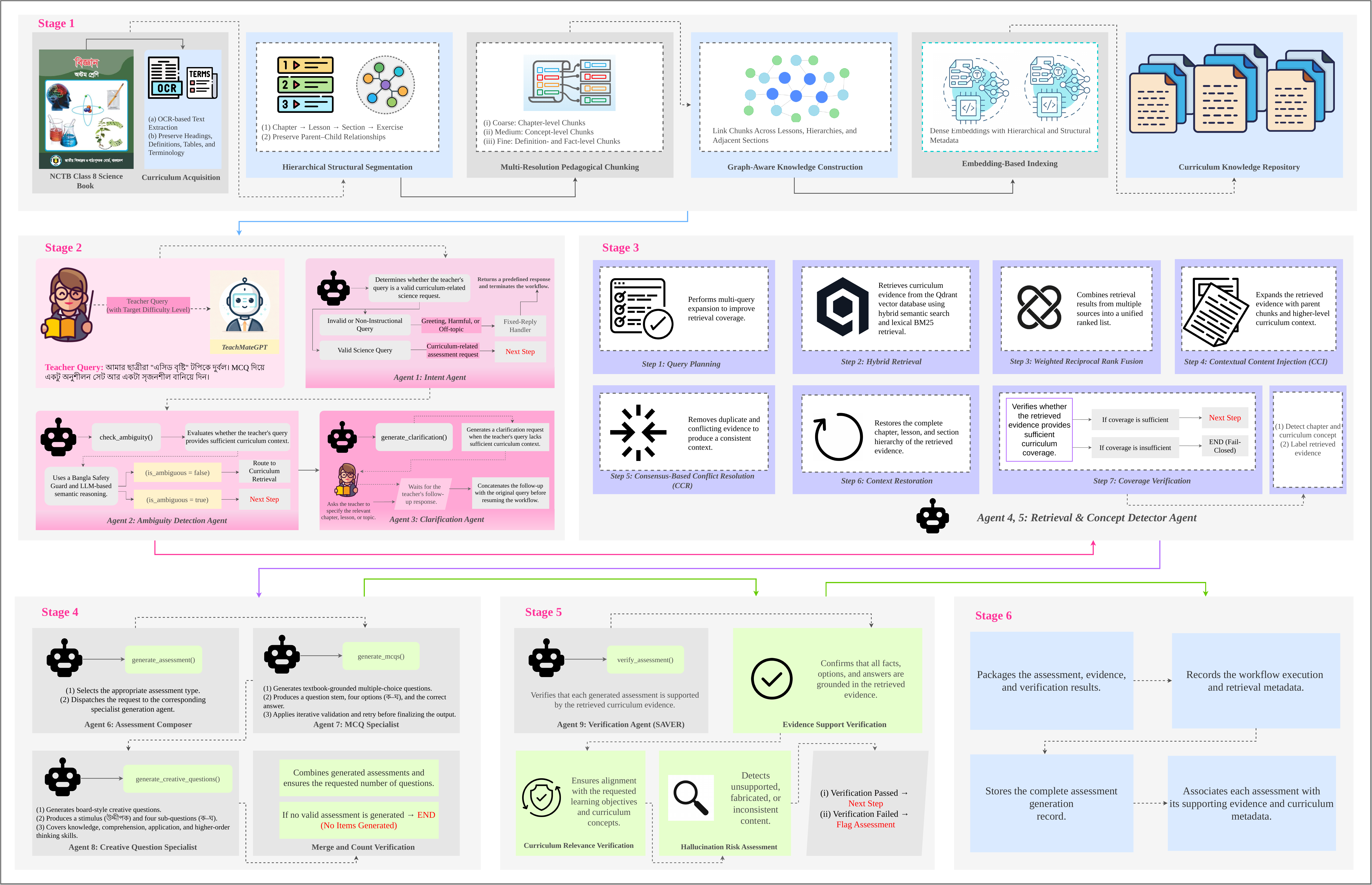}
\caption{End-to-end architecture of \textbf{\textsc{TeachMateGPT}.} Each teacher query flows through the COPE-based curriculum knowledge repository $\mathcal{K}$, intent analysis and routing, hybrid dense-BM25 retrieval with context restoration and coverage checks, generation of textbook-grounded MCQs and board-style creative questions ({\bns Ud/diipk}) with sub-questions {\bns k}--{\bns gh}), source-attributed verification of evidence support, faithfulness, and relevance, and packaging into an auditable, textbook-traceable output, as detailed in the following subsections.}
\label{fig:teachmategpt_framework}
\end{figure*}

Figure~\ref{fig:teachmategpt_framework} presents the end-to-end architecture of \textbf{\textsc{TeachMateGPT}}, which transforms a teacher's instructional request into curriculum-grounded assessments. Algorithm~\ref{alg:teachmategpt} (Appendix~\ref{algo}) summarizes the complete framework.

\subsection{Task Formulation}

Given a collection of authorized NCTB Class 8 science curriculum documents, the objective is to generate curriculum-aligned assessments supported by verifiable curriculum evidence.

Formally, let $\mathcal{D}={d_1,d_2,\ldots,d_n}$ denote the collection of curriculum documents. The framework first constructs a hierarchical curriculum knowledge repository,

\begin{equation}
\mathcal{K}=\mathrm{COPE}(\mathcal{D}),
\end{equation}

where \textbf{COPE} transforms curriculum documents into a hierarchical retrieval-ready representation.

Given a teacher query $q$, the retrieval module accesses the curriculum repository and returns the supporting evidence,

\begin{equation}
E = R(\mathcal{K}, q),
\end{equation}

where $R(\cdot)$ denotes the curriculum retrieval module.

Using the retrieved evidence $E$, the requested difficulty level $\ell$, and assessment type $S \in {\mathrm{MCQ}, \mathrm{CQ}}$, the generation module produces the assessment set,

\begin{equation}
\label{eq:generation}
A = G(E,\ell,S,q),
\end{equation}

where $G(\cdot)$ denotes the assessment generation module.

Finally, the generated assessments are verified by SAVER,

\begin{equation}
\label{eq:saver}
V = \mathrm{SAVER}(q,E,A),
\end{equation}

where $V$ is the verification report and $\mathrm{SAVER}(\cdot)$ performs source-attributed verification and evidence ranking, as detailed in Section~\ref{sec:saver}.

\subsection{Stage 1: Knowledge Base Construction (COPE)}
Reliable curriculum-grounded generation requires retrieval units that preserve the pedagogical structure of educational content. \textbf{COPE} constructs a hierarchical curriculum knowledge repository from authorized NCTB Class 8 science textbooks as a one-time preprocessing step before deployment.
COPE consists of five sequential steps: \textit{curriculum acquisition}, \textit{hierarchical structural segmentation}, \textit{multi-resolution pedagogical chunking}, \textit{graph-aware knowledge construction}, and \textit{embedding-based indexing}.
\paragraph{(1) Curriculum Acquisition.}
The framework collects authorized NCTB documents and converts them into machine-readable text while preserving chapter titles, section headings, definitions, examples, tables, question blocks, and scientific terminology across both digitally generated and scanned textbooks. Before segmentation, a normalization stage removes conversion and OCR artifacts while preserving semantic integrity. The result is a clean, structurally faithful corpus ready for hierarchical structural segmentation.

\paragraph{(2) Hierarchical Structural Segmentation.}
Unlike conventional RAG systems that use fixed-size windows, COPE models the instructional organization of curriculum documents directly. Each textbook is divided into structural units along pedagogical boundaries, chapters, lessons, sections, summaries, exercise blocks, and longer units are further decomposed into finer grains while preserving parent--child relationships. Every text segment thus retains its position in the curriculum hierarchy, letting retrieval exploit both local content and broader instructional context.

\paragraph{(3) Multi-Resolution Pedagogical Chunking.}
Educational queries vary in granularity: some require broad conceptual explanations, while others target specific definitions or factual details. COPE therefore represents curriculum content at multiple pedagogical resolutions rather than relying on a single fixed chunk size. Larger chunks preserve chapter-level context, intermediate chunks capture coherent concepts, and finer chunks isolate detailed knowledge, allowing retrieval to dynamically match the appropriate level of abstraction to the teacher's request. Table~\ref{tab:hierarchical_structure} summarizes the resulting chunk inventory and the windowing parameters used to construct each resolution tier.

\paragraph{(4) Graph-Aware Knowledge Construction.}
COPE further links chunks from the same lesson, hierarchy, or adjacent sections through a lightweight graph preserving sequential and hierarchical dependencies. This distinguishes COPE from a purely hierarchical parent-child index: rather than recording only a chunk's static parent, COPE also records sibling and sequential neighbor edges at ingest time, letting retrieval traverse this graph outward from a seed match to recover complementary evidence from related instructional units, rather than being limited to isolated passages or a single ancestor chunk. As Table~\ref{tab:research_gap} shows, curriculum-aware retrieval is at best partially supported among the compared Bangla NCTB systems; the graph-aware lineage introduced here lets COPE offer this capability in full, and its removal in the \textit{w/o COPE} ablation (Table~\ref{tab:automatic_evaluation}, Table~\ref{tab:human_evaluation}) accounts for the largest drop in answer relevancy among the five ablated components, alongside a substantial reduction in stimulus realism.

\paragraph{(5) Embedding-Based Curriculum Repository.}
Every chunk is encoded into a dense semantic representation together with pedagogical metadata (hierarchical level, structural position, graph relationships), forming the curriculum repository,

\begin{equation}
\mathcal{K}=\left\{(c,\mathbf{e}_c,\mathrm{meta}(c))\right\},
\end{equation}

where $c$ is a curriculum chunk, $\mathbf{e}_c$ its embedding, and $\mathrm{meta}(c)$ its hierarchical and structural information. $\mathcal{K}$ lets teacher queries access curriculum knowledge without repeating preprocessing or indexing. 

\subsection{Stage 2: Intent Analysis and Query Routing}

After constructing $\mathcal{K}$, a coordinated pipeline of three agents, the \textbf{Intent Agent}, \textbf{Ambiguity Detection Agent}, and \textbf{Clarification Agent}, collectively referred to as \textbf{IAC} (\textbf{I}ntent, \textbf{A}mbiguity-detection, and \textbf{C}larification), determines whether a teacher's request is curriculum-related before retrieval. The pipeline filters unrelated queries such as greetings, harmful requests, or off-topic inputs, identifies underspecified topics, and requests clarification before retrieval continues, ensuring that only well-specified requests proceed further.

\subsection{Stage 3: Hybrid Retrieval, Evidence Refinement, and Coverage Validation}

Routed requests enter a coordinated workflow of two agents, the \textbf{Retrieval Agent} and the \textbf{Concept Detector}, that transforms the refined query into a reliable evidence set. Rather than relying on a single semantic search, the Retrieval Agent internally decomposes retrieval into specialized steps: hierarchical hybrid retrieval, evidence enrichment (CCI), redundancy reduction (CCR), and evidence validation, ensuring generation proceeds only with sufficient curriculum evidence; the Concept Detector then labels the validated evidence with the curriculum chapter and concept it covers. Prompt specifications for the routing and generation agents are provided in Appendix~\ref{app:prompts}; the full implementation is available in the accompanying codebase.

\paragraph{(1) Hierarchical Hybrid Retrieval.}

The refined query goes to a hybrid retrieval module combining dense semantic retrieval, for related concepts, with lexical retrieval, for exact scientific terminology embeddings may miss. Rankings are merged via weighted reciprocal rank fusion,

\begin{equation}
\mathrm{Score}(c)=\sum_{r}\frac{w_r}{k+\mathrm{rank}_r(c)},
\end{equation}

where $w_r$ is the weight of retrieval method $r$, producing a ranked candidate set that balances semantic similarity with curriculum-specific lexical matching.

\paragraph{(2) Contextual Content Injection (CCI).}

Retrieved candidates may be isolated fragments of a larger concept. \textbf{CCI} restores pedagogical context by selectively incorporating higher-level instructional content tied to the retrieved segments, preserving conceptual continuity without indiscriminately expanding the evidence set.

\paragraph{(3) Consensus-Based Conflict Resolution (CCR).}

Multi-resolution retrieval produces overlapping evidence across hierarchical levels. \textbf{CCR} identifies equivalent or highly overlapping segments and retains the most informative representation, reducing redundancy while keeping complementary information for downstream reasoning. We use ``conflict'' here in the sense of overlapping or duplicated evidence spans competing for the same context budget, rather than evidence that reports contradictory facts; the implementation targets the former.

\paragraph{(4) Evidence Validation.}

This last check determines whether the evidence sufficiently supports the assessment task by examining curriculum coverage, structural diversity, and alignment with requested concepts. Let $T$ denote extracted curriculum concepts and $E$ the retrieved evidence set; coverage is

\begin{equation}
\mathrm{Cov}(T,E)=
\frac{\left|\left\{t\in T: t\in E\right\}\right|}
{|T|}.
\end{equation}

Generation proceeds only when coverage meets the required threshold; otherwise the framework withholds generation and requests a more specific query, preventing assessment authoring on insufficient evidence. Section~\ref{sec:results} and Table~\ref{tab:retrieval_ablation} report the resulting fail-closed rate and coverage ratio across retrieval configurations; we did not separately measure refusal precision or recall against a labeled ground truth of queries that should or should not have been refused, and note this as a scope limitation in Section~\ref{sec:limitations}.

\subsection{Stage 4: Multi-Format Assessment Generation}

Given validated evidence $E$, three agents generate classroom-ready assessments for the requested type and difficulty: the \textbf{Assessment Composer} dispatches the request to the \textbf{MCQ Specialist} and the \textbf{Creative Specialist}, which independently generate each format from the same evidence context. Rather than relying on a single prompt for all formats, each specialist follows its own pedagogical requirements while remaining grounded in the retrieved evidence.

\paragraph{(1) Evidence-Grounded Assessment Generation.}

Let $\ell$ be the requested difficulty and $S$ the desired type. The generation module produces the assessment set $A$ as defined in Equation~\eqref{eq:generation}. Since every generator receives the same validated evidence $E$, items stay consistent with retrieved curriculum content rather than relying on the model's parametric knowledge.

\paragraph{(2) Specialized Assessment Generation.}

The framework supports two formats used in Bangladeshi secondary education: MCQs and board-style CQs, operating on the same evidence while serving different pedagogical objectives. The MCQ generator produces a stem, four options, and one correct answer, grounding both the correct answer and distractors in the retrieved evidence for factual accuracy and curriculum alignment. The CQ generator constructs a contextual stimulus followed by four progressively structured sub-questions assessing knowledge, comprehension, application, and higher-order reasoning; rather than reproducing textbook passages, it synthesizes realistic scenarios consistent with the retrieved concepts.

\subsection{Stage 5: Source-Attributed Verification (SAVER)}
\label{sec:saver}

Curriculum-grounded retrieval reduces hallucination but does not eliminate it. Before presentation, every assessment passes through the \textbf{Verification Agent}, running \textbf{SAVER}, an independent post-generation verification process that compares each assessment against its retrieved evidence, as introduced in Equation~\eqref{eq:saver}, without modifying the generated content. This single agent then scores every assessment along three criteria.

\begin{enumerate}
    \item \textbf{\textit{Faithfulness.}} Whether facts, options, and statements are explicitly supported by the evidence.
    \item \textbf{\textit{Curriculum Relevance.}} Alignment between the assessment, the teacher's objective, and the retrieved concepts.
    \item \textbf{\textit{Hallucination Risk.}} Likelihood of unsupported, fabricated, or scientifically inconsistent content.
\end{enumerate}

These signals are aggregated into a report with quantitative scores and explanatory feedback, and are used to rank items by their evidence support so that the least-supported items surface first for teacher attention. An assessment is flagged for teacher attention rather than accepted for unedited release when

\begin{equation}
\small
\begin{aligned}
\mathrm{Accept}(V)=&
b_{\mathrm{faith}}
\land (s_{\mathrm{faith}}\geq\theta_F)\\
&\land (s_{\mathrm{rel}}\geq\theta_R)
\land (s_{\mathrm{hall}}\leq\theta_H)
\end{aligned}
\end{equation}

does not hold, where $b_{\mathrm{faith}}$ is the overall verification decision and $s_{\mathrm{faith}}$, $s_{\mathrm{rel}}$, $s_{\mathrm{hall}}$ are the faithfulness, relevance, and hallucination scores. Rather than regenerating or discarding failed assessments automatically, the framework preserves both the assessment and its verification report, so acceptance, editing, or discarding of flagged items remains a teacher decision. Export format and provenance details are given in Appendix~\ref{app:dataset_construction}.

\subsection{Stage 6: Auditable Output Packaging}
\label{sec:stage6_output_packaging}

The framework packages the teacher query, retrieved evidence with its provenance (source textbook, chapter, and concept labels), the generated assessment, the verification report, and an execution trace into a structured record. This record is presented to the teacher and also forms the basis of the \textbf{NCTB-SciGen8} dataset (Section~\ref{sec:dataset_construction}), ensuring that every dataset instance retains the same evidence trail available during generation.

\section{Dataset Construction}
\label{sec:dataset_construction}

\textbf{\textsc{TeachMateGPT}} also functions as the data-creation pipeline for \textbf{NCTB-SciGen8}, a reusable dataset of curriculum-grounded assessments assembled directly from verified pipeline outputs. Every assessment that passes \textbf{SAVER} verification (Stage~5, Section~\ref{sec:saver}) and output packaging (Stage~6, Section~\ref{sec:stage6_output_packaging}) becomes a dataset instance.

\textbf{NCTB-SciGen8} contains \textbf{198} Bangla Class 8 science assessments (\textbf{143} MCQs, \textbf{55} CQs) spanning all \textbf{14} chapters (\textbf{156} pages) of the official NCTB Class 8 Science textbook, each preserving its full generation provenance. Table~\ref{tab:kb_statistics} summarizes the per-chapter distribution of pages, subject areas, and assessment instances. The export schema, provenance format, teacher-reviewed subset, and coverage-based adequacy argument are detailed in Appendix~\ref{app:dataset_construction}; representative CQ and MCQ samples spanning multiple chapters are provided in Appendix~\ref{app:dataset_examples}.

\section{Experimental Setup}

The complete implementation details and evaluation settings used in our experiments are provided in Appendix~\ref{appendix:experimental_details}.

\section{Results Analysis}
\label{sec:results}

We evaluate \textbf{\textsc{TeachMateGPT}} through five research questions. Detailed analyses for each research question are provided in Appendix~\ref{Rq-appendix}. Automatic evaluation (Table~\ref{tab:automatic_evaluation}) and human evaluation (Table~\ref{tab:human_evaluation}) are presented below, whereas retrieval reliability under the fail-closed coverage gate (Table~\ref{tab:retrieval_ablation}) and inference-time and indexing efficiency (Table~\ref{tab:efficiency_comparison}) are reported in Appendix~\ref{app:ablation_study}.

\begin{table}[h]
\scriptsize
\centering
\setlength{\tabcolsep}{3pt}
\renewcommand{\arraystretch}{1.08}

\begin{tabular}{lcccc}
\toprule
\rowcolor{cyan!12}
\textbf{Configuration} &
\textbf{Faith.} $\uparrow$ &
\textbf{Ans. Rel.} $\uparrow$ &
\textbf{Ctx. Prec.} $\uparrow$ &
\textbf{Ctx. Rec.} $\uparrow$ \\
\midrule

Vanilla RAG 
& 0.68 & 0.60 & 0.54 & 0.58 \\

\rowcolor{orange!20}
\textbf{\textsc{TeachMateGPT}}
& \textbf{\textcolor{DarkGreen}{0.96}}
& \textbf{\textcolor{DarkGreen}{0.89}}
& \textbf{\textcolor{DarkGreen}{0.92}}
& \textbf{\textcolor{DarkGreen}{0.91}} \\

w/o COPE
& 0.86 & 0.76 & 0.73 & 0.75 \\

w/o SAVER
& 0.79 & 0.86 & 0.90 & 0.89 \\

w/o CCR
& 0.83 & 0.81 & 0.84 & 0.85 \\

w/o CCI
& 0.84 & 0.82 & 0.85 & 0.86 \\

w/o IAC
& 0.81 & 0.77 & 0.89 & 0.88 \\

RAPTOR
& 0.80 & 0.75 & 0.71 & 0.78 \\

GraphRAG
& 0.84 & 0.79 & 0.80 & 0.83 \\

CRAG
& 0.86 & 0.81 & 0.84 & 0.85 \\

Adaptive RAG
& 0.79 & 0.76 & 0.73 & 0.79 \\

\bottomrule
\end{tabular}

\caption{RAGAS evaluation of \textbf{\textsc{TeachMateGPT}} against a vanilla RAG baseline, four RAG baselines, and five component ablations (w/o COPE, SAVER, CCR, CCI, IAC). Faith. = Faithfulness, Ans. Rel. = Answer Relevancy, Ctx. Prec. = Context Precision, Ctx. Rec. = Context Recall; $\uparrow$ indicates higher is better. It attains the best score on all four metrics.}
\label{tab:automatic_evaluation}
\end{table}

\begin{table*}[h]
\small
\setlength{\tabcolsep}{3pt}
\renewcommand{\arraystretch}{1.08}
\centering

\begin{tabular}{lcccc}
\toprule
\rowcolor{cyan!12}
\textbf{Configuration} &
\textbf{Pedagogical Alignment} $\uparrow$ &
\textbf{Stimulus Realism} $\uparrow$ &
\textbf{Linguistic Fluency} $\uparrow$ &
\textbf{Overall Utility} $\uparrow$ \\
\midrule

Vanilla RAG &
2.40 $\pm$ 0.12 &
1.75 $\pm$ 0.10 &
3.00 $\pm$ 0.12 &
2.00 $\pm$ 0.10 \\

\rowcolor{orange!20}
\textbf{\textsc{TeachMateGPT}} &
\textbf{\textcolor{DarkGreen}{4.90 $\pm$ 0.10}} &
\textbf{\textcolor{DarkGreen}{4.70 $\pm$ 0.10}} &
\textbf{\textcolor{DarkGreen}{4.60 $\pm$ 0.10}} &
\textbf{\textcolor{DarkGreen}{4.80 $\pm$ 0.10}} \\

w/o COPE &
4.05 $\pm$ 0.10 &
3.90 $\pm$ 0.12 &
4.35 $\pm$ 0.08 &
3.95 $\pm$ 0.10 \\

w/o SAVER &
3.25 $\pm$ 0.12 &
4.25 $\pm$ 0.10 &
4.25 $\pm$ 0.10 &
3.55 $\pm$ 0.12 \\

w/o CCR &
3.80 $\pm$ 0.10 &
4.05 $\pm$ 0.10 &
4.30 $\pm$ 0.08 &
3.95 $\pm$ 0.10 \\

w/o CCI &
3.75 $\pm$ 0.10 &
4.00 $\pm$ 0.10 &
4.25 $\pm$ 0.10 &
3.85 $\pm$ 0.10 \\

w/o IAC &
3.50 $\pm$ 0.12 &
3.85 $\pm$ 0.12 &
4.15 $\pm$ 0.10 &
3.60 $\pm$ 0.12 \\

RAPTOR &
3.55 $\pm$ 0.12 &
3.60 $\pm$ 0.12 &
4.00 $\pm$ 0.10 &
3.55 $\pm$ 0.12 \\

GraphRAG &
3.70 $\pm$ 0.10 &
3.85 $\pm$ 0.10 &
4.10 $\pm$ 0.10 &
3.75 $\pm$ 0.10 \\

CRAG &
3.85 $\pm$ 0.10 &
3.95 $\pm$ 0.10 &
4.20 $\pm$ 0.08 &
3.95 $\pm$ 0.10 \\

Adaptive RAG &
3.60 $\pm$ 0.10 &
3.75 $\pm$ 0.10 &
4.05 $\pm$ 0.10 &
3.70 $\pm$ 0.10 \\

\bottomrule
\end{tabular}
\caption{Human evaluation of \textbf{\textsc{TeachMateGPT}}-generated Bangla Class~8 NCTB science assessment items by three practicing science teachers. Teachers rated each item on a 5-point Likert scale across four criteria: Pedagogical Alignment, Stimulus Realism, Linguistic Fluency, and Overall Utility. Values denote mean $\pm$ standard deviation across generated samples and reflect run-to-run variation in model outputs; $\uparrow$ indicates higher is better. We compare the complete pipeline with five component-level ablations and four representative RAG baselines. \textbf{\textsc{TeachMateGPT}} achieves the highest score for all four evaluation criteria.}
\label{tab:human_evaluation}
\end{table*}

\textbf{RQ1: Effectiveness of COPE in Curriculum Knowledge Base Construction.}
COPE's hierarchical index preserves the NCTB curriculum structure across all 14 chapters with balanced depth across subject areas. Four of six deterministic validity gates achieve 100\% pass rate (Table~\ref{tab:validation_gates}). The remaining errors relate to formatting constraints (option format and scientific notation), not missing curriculum evidence, showing that the curriculum representation layer provides sufficient grounding for assessment generation.

\textbf{RQ2: Performance of the Intent and Clarification Routing Layer.}
The routing layer removes 30\% of evaluation queries before retrieval, including greetings, harmful requests, and off-topic inputs (Table~\ref{tab:intent_routing}). The Bangla specificity guard resolves 81\% of ambiguity cases without model intervention, and explicit teacher prompts trigger no unnecessary clarification (Table~\ref{tab:ambiguity_gate}). Thus, lightweight routing improves safety + efficiency while maintaining usability.

\textbf{RQ3: Reliability of Hybrid Retrieval and the Fail-Closed Coverage Gate.}
Hybrid retrieval with a coverage gate achieves a safety--coverage balance: 12.5\% fail-closed rate with 0.724 coverage ratio (Table~\ref{tab:retrieval_ablation}). Dense-only retrieval increases refusal to 31.3\% and lowers coverage to 0.618, while gate removal reduces safety despite fewer refusals. These results show that dense + lexical retrieval provide complementary signals for OCR-affected Bangla curriculum text.

\textbf{RQ4: Quality of Curriculum-Grounded Assessment Generation.}
MCQ generation achieves 85.7\% first-attempt validation success, while CQ generation rises from 7.1\% → 100\% after CQ narrative adjustment (Table~\ref{tab:generation_yield}). This result indicates that initial CQ errors mainly came from narrative-style mismatch rather than weak curriculum grounding. MCQ stem length remains nearly unchanged across difficulty levels (Table~\ref{tab:difficulty_effect}), suggesting that difficulty depends on semantic and reasoning factors rather than surface length.

\textbf{RQ5: Validation of Source-Attributed Verification and Teacher Review.}
SAVER identifies only structural defects, with no fabricated facts detected in the audited sample (all 55 CQ clues and a 15-item MCQ spot check; Table~\ref{tab:saver_analysis}). Compared with Vanilla RAG, \textbf{\textsc{TeachMateGPT}} improves faithfulness ($0.68 \uparrow 0.96$), context precision ($0.54 \uparrow 0.92$), and teacher utility ($2.00 \uparrow 4.80$) (Tables~\ref{tab:automatic_evaluation} and~\ref{tab:human_evaluation}). Ablations show distinct roles: removing COPE mainly reduces retrieval quality and stimulus realism, while removing SAVER causes the largest drop in faithfulness and pedagogical alignment.

\section{Conclusion}

We introduce \textbf{\textsc{TeachMateGPT}}, a curriculum-grounded multi-agent framework for Bangla assessment generation. Our contributions are fourfold: \textbf{(1)} \textbf{COPE} (\textbf{C}urriculum-\textbf{O}riented \textbf{P}edagogical \textbf{E}mbedding), a hierarchical, graph-aware curriculum index whose ingest-time lineage and neighbor links let retrieval exceed static parent-child chunks; \textbf{(2)} a staged, fail-closed multi-agent pipeline that withholds generation under insufficient evidence rather than returning fabricated assessments; \textbf{(3)} \textbf{SAVER} (\textbf{S}ource-\textbf{A}ttributed \textbf{V}erification and \textbf{E}vidence \textbf{R}anking), a verification layer that scores faithfulness, relevance, and hallucination risk and flags unsupported items for teacher review; and \textbf{(4)} \textbf{NCTB-SciGen8}, a curriculum-grounded NCTB Class 8 science assessment dataset with a teacher-rated subset reviewed by three teachers. Across automatic and human evaluations, \textbf{\textsc{TeachMateGPT}} improves context precision from \textbf{0.54} to \textbf{0.92} and context recall from \textbf{0.58} to \textbf{0.91}. Ablation results further demonstrate the complementary roles of retrieval and verification: removing \textbf{COPE} reduces answer relevancy \textbf{0.89 $\downarrow$ 0.76} and pedagogical alignment \textbf{4.90 $\downarrow$ 4.05}, while removing \textbf{SAVER} causes the largest drop in faithfulness \textbf{0.96 $\downarrow$ 0.79} and pedagogical alignment \textbf{4.90 $\downarrow$ 3.25}. These findings show that trustworthy assessment generation depends on reliable retrieval, evidence-grounded verification, and an appropriate refusal to generate when curriculum evidence is insufficient. Although our study focuses on the Bangla NCTB Class 8 science textbook, \textbf{\textsc{TeachMateGPT}} provides a foundation for curriculum-grounded assessment generation. Future work will explore adaptation across curricula and languages, verification-guided refinement, and psychometric calibration of generated assessments.

\section*{Limitations}
\label{sec:limitations}

\textbf{Scope and Transferability.} \textbf{\textsc{TeachMateGPT}} focuses on Bangla Class 8 science assessment generation from authorized NCTB textbooks. We do not claim that the framework transfers directly to other grades, subjects, languages, or curricula. Several components, such as the Bangla specificity guard, board-style CQ constraints, and curriculum heading detectors, are specific to the NCTB curriculum. Extending the framework to new educational settings would therefore require index reconstruction and pipeline adaptation.

\textbf{Indexing and Curriculum Representation.} COPE relies on native text extraction or vision-based transcription of textbook pages, both of which may introduce OCR errors, incomplete page coverage, or corrupted mathematical notation. Because retrieval follows a fail-closed design, such errors lead to refusal or limited evidence rather than unsupported generation, although the resulting loss in recall remains only partially quantified. In addition, COPE captures structural relationships within the textbook rather than an explicit prerequisite or learning-objective graph, which may omit pedagogically related content outside the local textbook structure.

\textbf{Text-Only Modality and Diagram-Dependent Items.} \textbf{\textsc{TeachMateGPT}} is a text-only framework: retrieval, generation, and verification all operate over transcribed textbook prose, and vision is confined to the ingestion stage, where scanned pages are converted into text. Curriculum figures, such as circuit and ray diagrams, microscopic cell and organism illustrations, atomic-structure schematics, and labeled graphs, are consequently collapsed into text or discarded rather than retained as retrievable or reproducible visual objects. The framework therefore cannot author items whose stimulus or stem is itself a figure, for instance an MCQ that requires reading a given circuit or a creative-question {\bng Ud/diipk} organized around a diagram (``{\bng inecr ictRiT lkK kr}''). Such figure-dependent items are standard in NCTB board examinations, particularly for chapters including Circuit and Current Electricity, Light, and Structure of the Atom, so both the generated items and the released NCTB-SciGen8 dataset are biased toward verbal reasoning and under-represent this component of the curriculum. Extending COPE to multimodal indexing and figure-conditioned generation is a direction we leave to future work.

\textbf{Retrieval, Generation, and Verification.} The fail-closed retrieval strategy improves evidence quality but reduces recall by rejecting partially relevant evidence under paraphrases, synonymy, or OCR-induced lexical mismatch. We characterize this behavior only through the fail-closed rate and mean coverage ratio measured across retrieval ablations (Table~\ref{tab:retrieval_ablation}); we did not construct a labeled set of queries with gold refusal decisions, so refusal precision, refusal recall, and false-refusal rate against such a ground truth remain unmeasured, and the fail-closed and coverage figures we report should be read as descriptive of pipeline behavior on our evaluation bank rather than as calibrated detection metrics. Clarification also depends on teacher responses, so underspecified single-turn requests terminate without assessment generation. Generation quality remains bounded by the capabilities of the underlying language models. CQ quality is sensitive to narrative style, while force-filled outputs after validation failure may be pedagogically weaker than fully generated responses. Difficulty control relies on prompting rather than psychometric calibration, and SAVER identifies unsupported or low-confidence items but does not automatically revise or remove them from the released assessment.

\textbf{Evaluation Scale.} Our human evaluation relies on three practicing teachers rating a configuration-blind sample, and several component analyses use correspondingly small query sets. This limited evaluator pool and sample size constrain statistical power and inter-rater generalizability, so the reported ratings and agreement should be read as indicative rather than definitive; larger teacher panels and evaluation banks are needed to establish agreement and effect sizes more robustly.

\section*{Ethical Considerations}

\textbf{Intended Use and Human Oversight.} \textbf{\textsc{TeachMateGPT}} is designed as an assistive drafting tool for teachers, not as an autonomous assessment authority. Every generated item is presented together with its supporting evidence and its verification report, and the framework warns rather than silently rewriting flagged items, so a qualified teacher makes the final decision to accept, edit, or discard each assessment before classroom use. We caution against deploying the system in a fully automated setting, for example generating live examinations without human review, because automation bias may lead users to over-trust fluent but subtly incorrect items. Assessments produced by the system should be labeled as AI-assisted so that teachers, students, and reviewers remain aware of their origin.

\textbf{Curriculum Data and Copyright.} All curriculum content is drawn exclusively from the officially authorized NCTB Class~8 science textbook, a publicly distributed national curriculum resource, and we deliberately exclude third-party notes, commercial question banks, and unrestricted web material. The textbook remains the intellectual property of the National Curriculum and Textbook Board of Bangladesh; we use it for non-commercial research and do not redistribute the textbook itself. The NCTB-SciGen8 records store source identifiers such as chapter and section labels, and where a supporting passage is included it is limited to a short excerpt of at most one to two sentences retained solely for evidence traceability; we do not redistribute textbook pages or the textbook in full. To preserve anonymity during review, the dataset is not distributed with this submission. The dataset will be made publicly available under the CC BY-NC 4.0 license for non-commercial research use.

\textbf{Human Evaluation and Participant Treatment.} Our human evaluation involves three practicing secondary-school science teachers who rated a configuration-blind sample of generated assessments. The teachers are practicing educators who participated voluntarily and gave informed consent, without monetary compensation. No students or other minors took part in the study. The evaluation collected only pedagogical quality judgments about the generated items and no personal, sensitive, or identifying data about the teachers or any third party, and it posed minimal risk. All ratings are reported in aggregate. 

\textbf{Reliability, Misuse, and Academic Integrity.} Because incorrect assessment items could mislead learners or reinforce misconceptions, factual reliability is a central ethical concern. We mitigate this risk through curriculum-grounded retrieval, a fail-closed coverage gate that refuses generation under weak evidence, and a post-generation verification step that scores faithfulness, relevance, and hallucination risk against the retrieved evidence. These safeguards reduce rather than eliminate error, so teacher review remains necessary before any item reaches students. We also acknowledge the risk that a generation tool of this kind could be misused, for instance to mass-produce low-quality question banks or to circumvent a teacher's own assessment design, and we therefore position the system as support for, rather than replacement of, professional pedagogical judgment.

\textbf{Safety for a Minors-Adjacent Audience.} Because the system serves an educational context that includes school-age learners, an intent-routing stage screens every input before retrieval or generation. Unsafe or inappropriate requests, including violence, self-harm, weapons, sexual content involving minors, harassment, and cheating assistance, receive a fixed safe response instead. This routing is a first-line safeguard rather than a complete content-moderation guarantee, and teacher oversight remains part of safe deployment.

\textbf{Bias, Fairness, and Language.} The underlying language models may encode social and topical biases, and generated Bangla text can contain fluency or terminology errors that are harder to detect automatically in a low-resource language than in English. Difficulty labels are conveyed through prompting rather than psychometric calibration and should not be interpreted as validated measures of item difficulty. At the same time, by targeting Bangla NCTB science, this work aims to broaden access to assessment-authoring support for an underserved language community. We encourage similarly careful, curriculum-grounded, and human-supervised adaptation before the framework is extended to other languages, curricula, or learner populations.

\bibliography{custom}

\appendix

\section*{Appendix}
\section{Dataset Construction Details}
\label{app:dataset_construction}

This appendix provides the dataset export schema and the adequacy argument for \textbf{NCTB-SciGen8} that were summarized in Section~\ref{sec:dataset_construction} of the main text. The full \textbf{SAVER} formalization is presented in Section~\ref{sec:saver}.

\subsection{Export Format and Provenance}

Generated assessments are exported with full provenance that records the original query, detected chapter and concept, retrieved textbook sources, generated content, and verification results. MCQs follow the NCTB format with Bangla stems and four options labeled {\bns k}, {\bns kh}, {\bns g}, and {\bns gh}. CQs follow the board-style structure with an \emph{Uddipok} stimulus and four cognitive-level components: \emph{Gyan} (knowledge), \emph{Onudhabon} (comprehension), \emph{Proyog} (application), and \emph{Ucchotor Dokkhota} (higher-order skills). Unlike manually authored evaluation sets, items are retained only after passing retrieval and validation gates, and every retained item carries the SAVER verification report (Section~\ref{sec:saver}), so every released item is accompanied by an evidence trail rather than being filtered by SAVER's binary decision alone.

\subsection{Dataset Schema and Teacher Review}

All assessment instances are stored in JSON, each uniquely identified by \texttt{question\_id}. The schema preserves teacher requests, curriculum metadata, retrieved evidence, generated outputs, and evaluation annotations. Table~\ref{tab:dataset_schema} provides an overview of the schema.

\subsection{Adequacy as an Evaluation Dataset}

We position \textbf{NCTB-SciGen8} as a curated evaluation dataset rather than a large item bank, and its adequacy depends on coverage and quality rather than raw size. First, the 198 assessment items span all 14 chapters of the NCTB Class 8 science syllabus, providing complete curricular coverage of this bounded domain rather than a partial sample. Second, every item passes the validation gates before release (Table~\ref{tab:validation_gates}), and a subset is independently reviewed by three practicing science teachers (Section~\ref{sec:evaluation}), providing both automatic grounding checks and expert pedagogical judgment. Third, every instance preserves its generation provenance, retrieved evidence, chapter and concept labels, and verification report, enabling item-level auditing. We therefore consider NCTB-SciGen8 adequate for evaluating curriculum-grounded generation within this domain, while acknowledging, as discussed in Section~\ref{sec:limitations}, that larger datasets and broader evaluation panels are needed for more generalizable conclusions.

\section{\textbf{NCTB-SciGen8} Dataset Details}
\label{app:dataset_examples}

\begin{figure*}[h]
\centerline{\includegraphics[width=\textwidth]{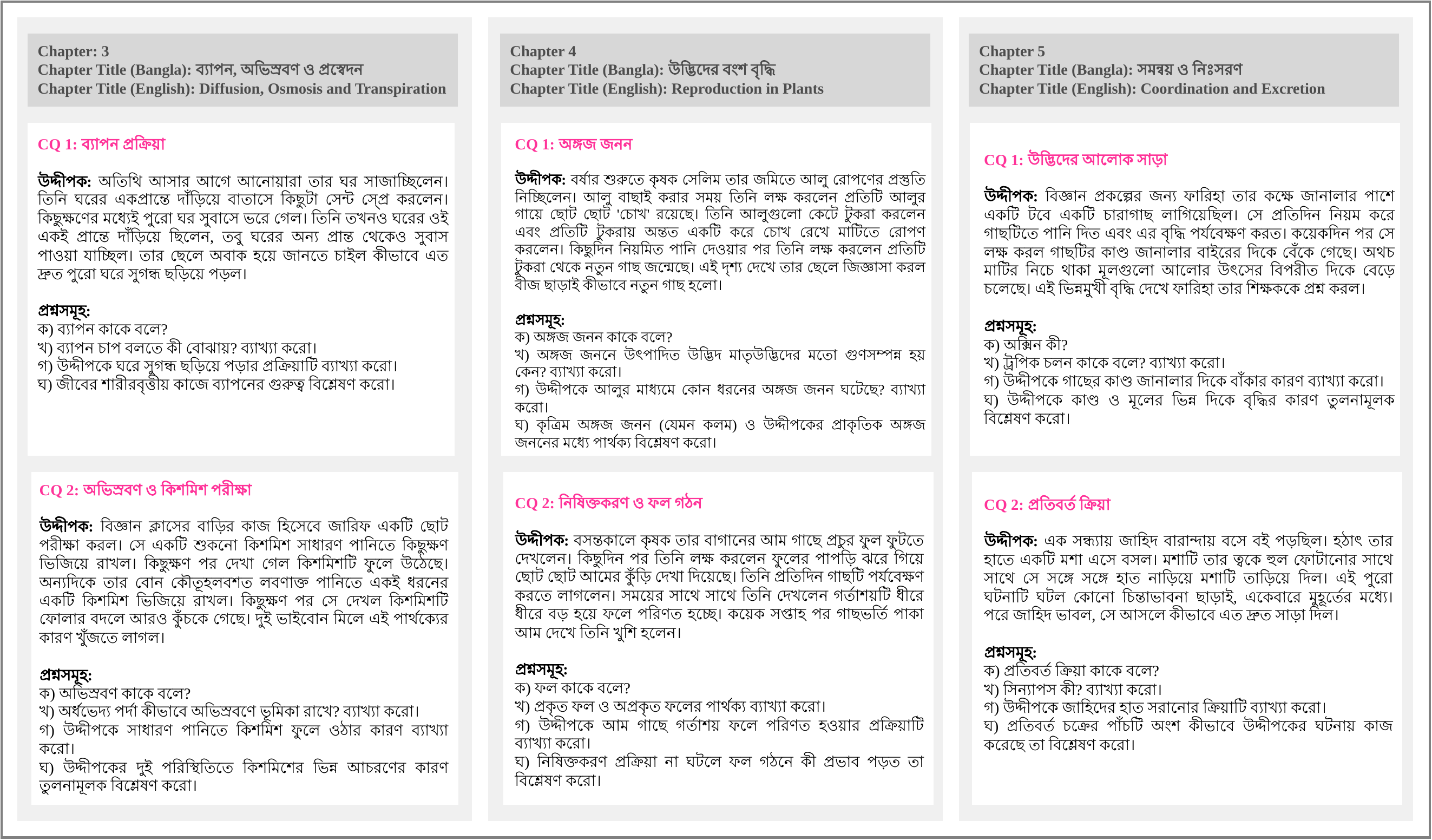}}
\caption{Examples of \textbf{chapter-wise creative questions} generated by \textbf{\textsc{TeachMateGPT}} for Chapter~3 (\textit{Diffusion, Osmosis and Transpiration}), Chapter~4 (\textit{Reproduction in Plants}), and Chapter~5 (\textit{Coordination and Excretion}) of the NCTB Class~8 Science textbook. For each chapter, TeachMateGPT retrieves chapter-specific curriculum evidence and produces a board-style {\bns Ud/diipk} with four progressive sub-questions ({\bns k}--{\bns gh}). The examples show consistent curriculum grounding, board-style structure, and formatting across different science topics.}
\label{fig:chapter_examples}
\end{figure*}

\begin{figure*}[h]
\centerline{\includegraphics[width=\textwidth]{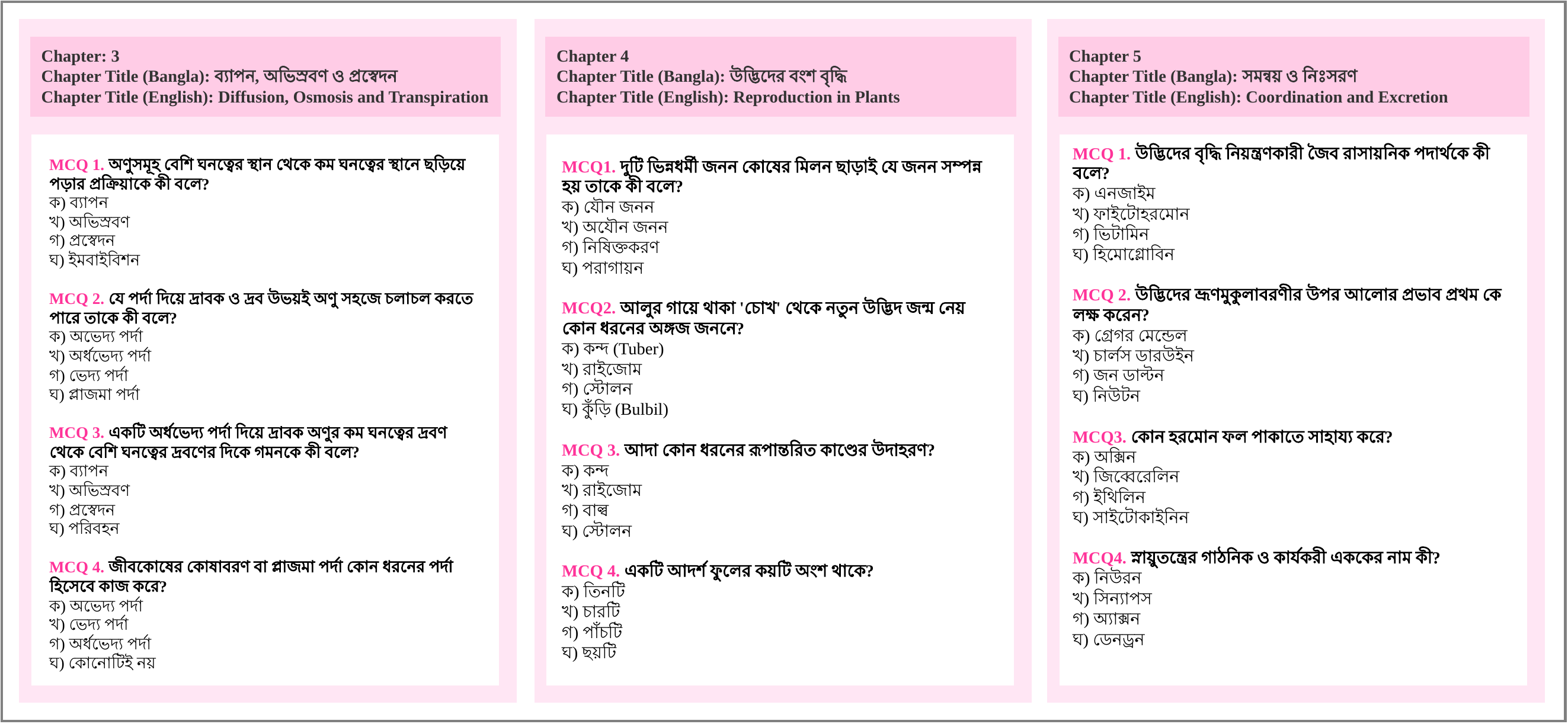}}
\caption{Examples of \textbf{MCQ items} from Chapter~3 (\textit{Diffusion, Osmosis and Transpiration}), Chapter~4 (\textit{Reproduction in Plants}), and Chapter~5 (\textit{Coordination and Excretion}) of the NCTB Class~8 Science textbook illustrate how the \textbf{\textsc{TeachMateGPT}} framework converts the same retrieved evidence used for creative-question generation into a compact, single-answer format. For each chapter, the MCQ specialist grounds a Bangla stem in the source passage and generates four options ({\bns k}--{\bns gh}), from which the student selects the correct answer, while the remaining options serve as plausible, textbook-consistent distractors rather than invented facts.}
\label{fig:mcq_chapter_examples}
\end{figure*}

\begin{figure*}[h]
\centering
\begin{subfigure}{0.48\textwidth}
    \centering
    \includegraphics[width=\linewidth]{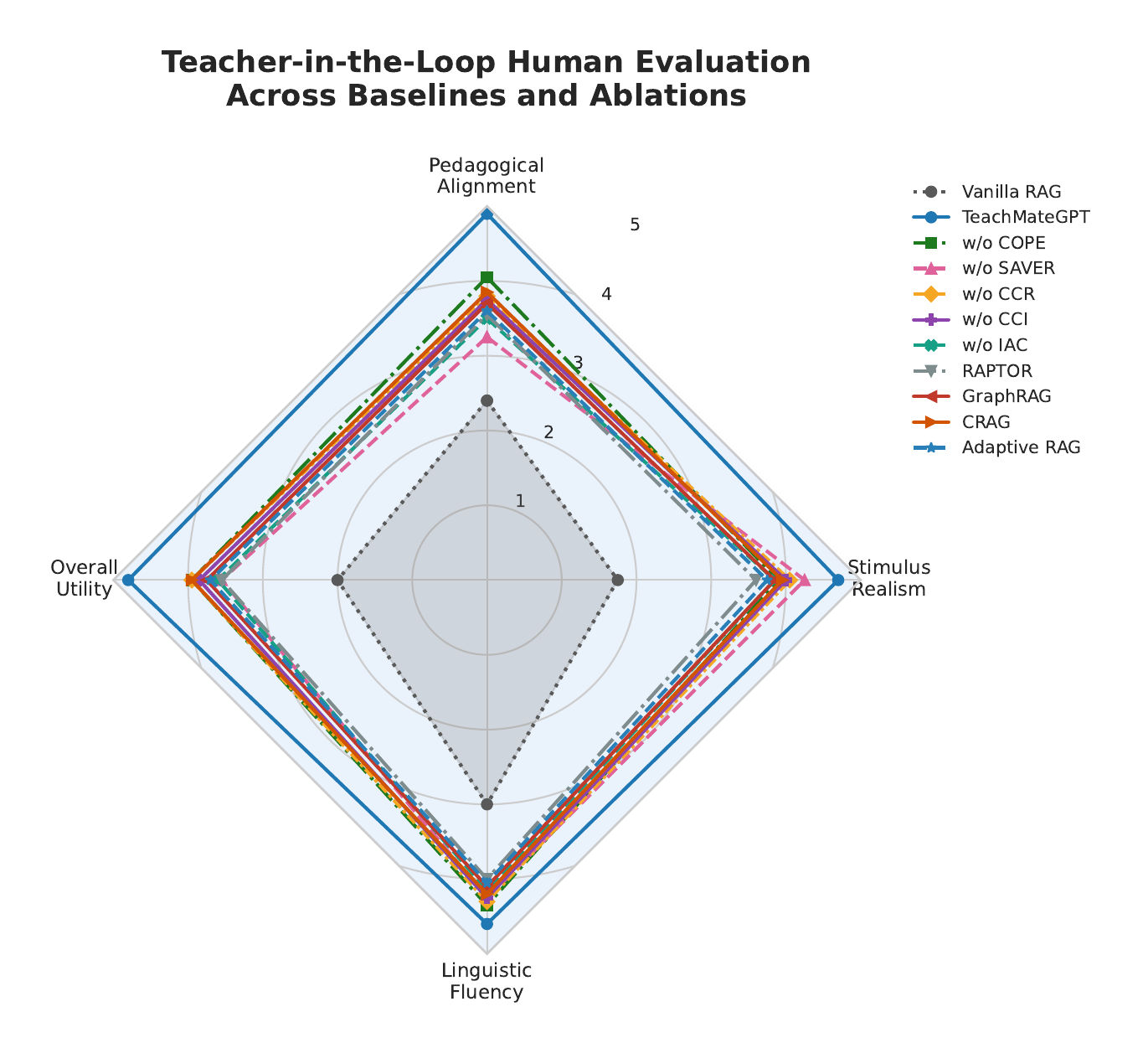}
    \caption{Teacher-in-the-loop evaluation.}
    \label{fig:teacher_eval_radar}
\end{subfigure}
\hfill
\begin{subfigure}{0.48\textwidth}
    \centering
    \includegraphics[width=\linewidth]{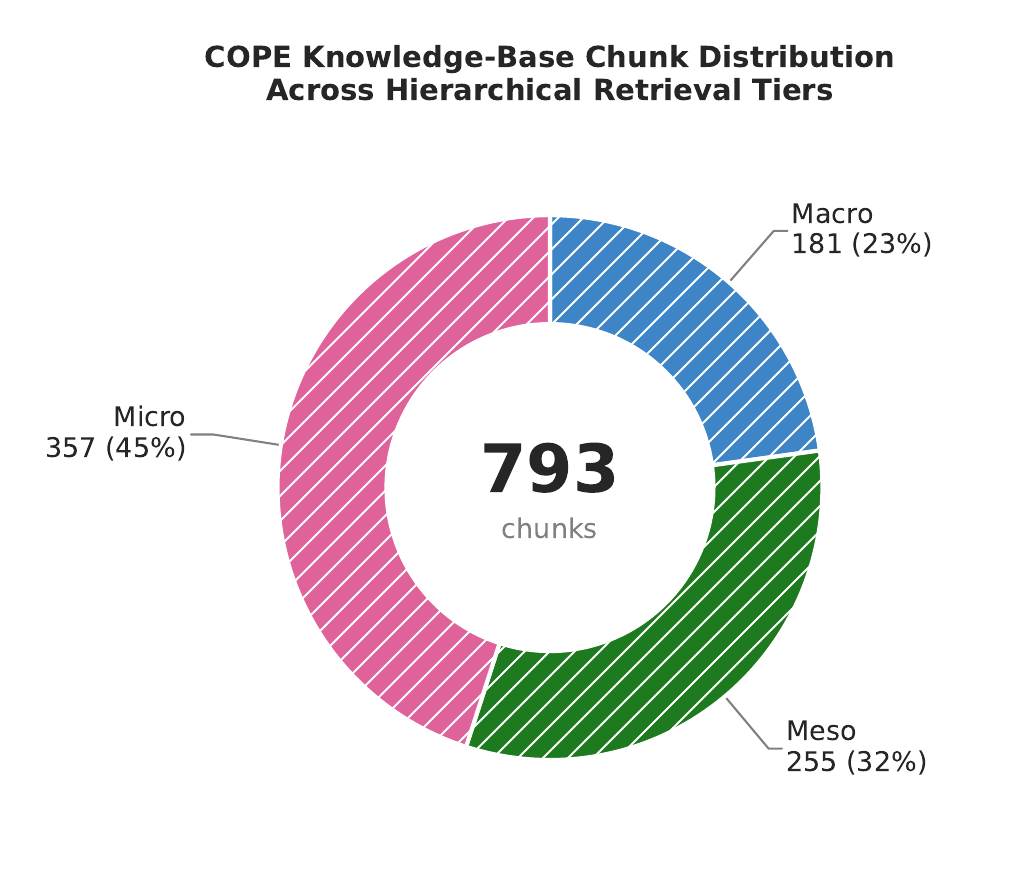}
    \caption{COPE chunk tier composition.}
    \label{fig:chunk_tier_distribution}
\end{subfigure}
\caption{Qualitative and structural analysis of \textbf{\textsc{TeachMateGPT}}. \textbf{(a)} Human evaluation by three science teachers across four assessment-quality criteria, showing that the full system consistently outperforms all ablations. \textbf{(b)} Distribution of the hierarchical knowledge base across the three COPE resolution tiers, illustrating the multi-resolution chunking strategy for broad contextual and fine-grained factual retrieval.}
\label{fig:teacher_chunk_analysis}
\end{figure*}

\begin{figure*}[h]
\centerline{\includegraphics[width=\textwidth]{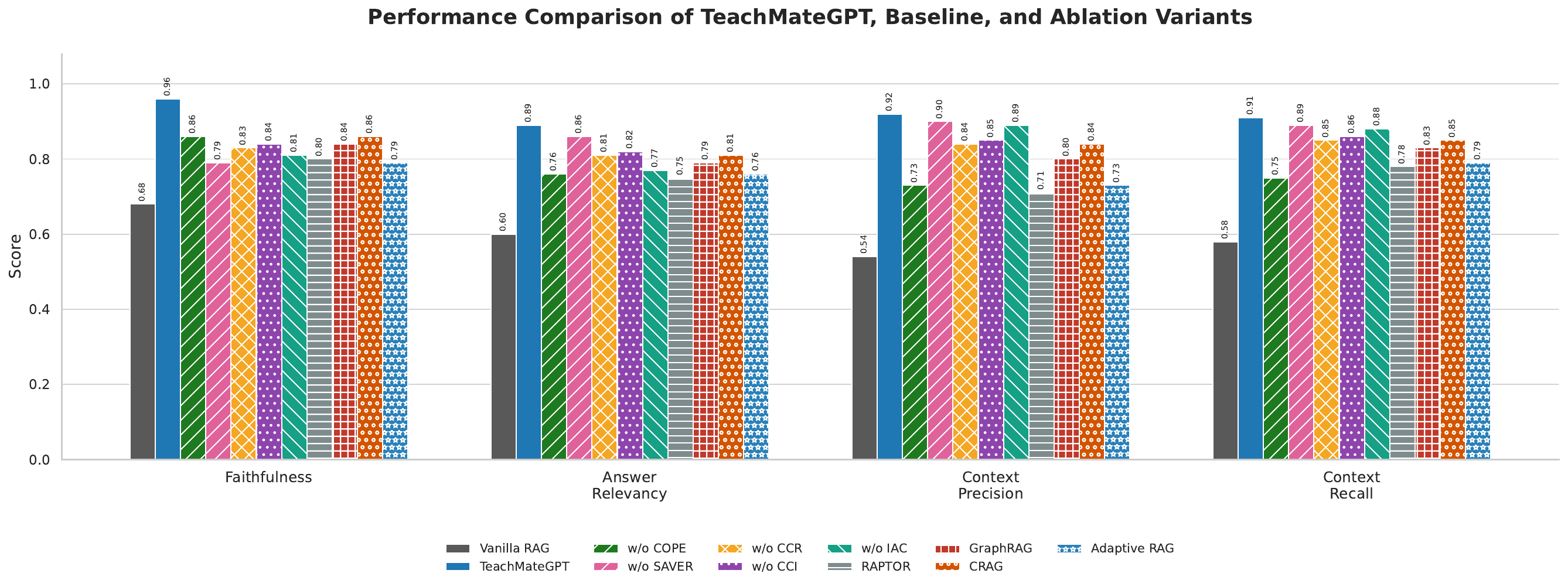}}
\caption{Quantitative comparison of \textbf{\textsc{TeachMateGPT}} with a vanilla RAG baseline, five component ablations (w/o COPE, SAVER, CCR, CCI, and IAC), and three representative RAG baselines (RAPTOR, GraphRAG, and CRAG) across four generation metrics: Faithfulness, Answer Relevancy, Context Precision, and Context Recall (Table~\ref{tab:automatic_evaluation}). Each ablation removes one architectural component to measure its contribution to retrieval quality and assessment generation.}
\label{fig:ragas_ablation_bar}
\end{figure*}

Figures~\ref{fig:chapter_examples} and~\ref{fig:mcq_chapter_examples} present sample CQ and MCQ examples from \textbf{NCTB-SciGen8}, while Tables~\ref{tab:hierarchical_structure} and~\ref{tab:kb_statistics} summarize the COPE indexing configuration and the chapter-wise dataset statistics, respectively. Figure~\ref{fig:ragas_ablation_bar} visualizes the automatic-evaluation ablations from Table~\ref{tab:automatic_evaluation}, and Figure~\ref{fig:teacher_chunk_analysis} presents the corresponding human-evaluation comparison alongside the COPE chunk-tier composition.

% TABLE 3 - COPE Chunk Inventory and Chunking Windows
% ============================================================
\begin{table}[h]
\setlength{\tabcolsep}{3pt}
\renewcommand{\arraystretch}{1.05}
\scriptsize

\centering

\setlength{\tabcolsep}{8pt}
\renewcommand{\arraystretch}{1.15}

\begin{tabular}{l r}
\toprule
\rowcolor{cyan!12}
\textbf{Quantity} & \textbf{Value} \\
\midrule
Chapters ingested & 14 / 14 \\
Total chunks & 793 \\
Macro-tier chunks & 181 \\
Meso-tier chunks & 255 \\
Micro-tier chunks & 357 \\
Token-aligned micro tier & disabled \\
Embedded tiers & macro, meso, micro \\
Macro-tier window (chunk size / overlap, chars) & 3600 / 650 \\
Meso-tier window (chunk size / overlap, chars) & 1900 / 320 \\
Micro-tier window (chunk size / overlap, chars) & 1100 / 180 \\
Token-micro window (disabled; tokens) & 640 / 96 \\
\bottomrule
\end{tabular}
\caption{Chunk counts and windowing parameters for the three nested COPE resolution tiers (macro, meso, micro) produced from 14 ingested NCTB Class 8 science chapters. Each tier is formed by recursively re-segmenting previous tier spans at progressively finer character windows (with overlap) to support broad conceptual retrieval and localized fact retrieval; the token-aligned micro variant is disabled by default due to OCR noise amplification.}
\label{tab:hierarchical_structure}
\end{table}

% ============================================================
% TABLE 4 - Chapter Inventory and Exam-Depth Capacity
% ============================================================
\begin{table*}[h]
\setlength{\tabcolsep}{3pt}
\renewcommand{\arraystretch}{1.05}
\scriptsize
\centering

\begin{tabular}{c l l c r r l}
\toprule
\rowcolor{cyan!12}
\textbf{Ch.} & \textbf{Chapter Name (Bangla)} & \textbf{Chapter Name (English)} & \textbf{Pages} & \textbf{MCQ} & \textbf{CQ} & \textbf{Subject} \\
\midrule
1  & {\bng pRaiNjgetr eshRiNibnYas}    & Classification of Animal Kingdom     & 1--12    & 13  & 4  & Biology \\
2  & {\bng jiiebr brRid/dh {O} bNNGshgit}        & Growth \& Heredity of Organisms      & 13--23   & 11  & 4  & Biology \\
3  & {\bng bYapn, AibhsRbN {O} pResWdn} & Diffusion, Osmosis \& Transpiration  & 24--33   & 10  & 4  & Biology \\
4  & {\bng Uid/bhedr bNNGsh brRid/dh}          & Reproduction in Plants               & 34--44   & 10  & 4  & Biology \\
5  & {\bng smnWJ {O} in{h}srN}              & Coordination \& Excretion            & 45--54   & 9   & 4  & Biology \\
6  & {\bng prmaNur gThn}                  & Structure of the Atom                & 55--64   & 11  & 4  & Chemistry \\
7  & {\bng prRithbii {O} mHakr/Sh}             & Earth \& Gravitation                 & 65--74   & 10  & 4  & Physics \\
8  & {\bng rasaJink ibikRJa}           & Chemical Reactions                   & 75--88   & 12  & 5  & Chemistry \\
9  & {\bng br/tnii {O} clibdYut//}            & Circuit \& Current Electricity       & 89--97   & 10  & 4  & Physics \\
10 & {\bng AmL, kKark {O} lbN}            & Acid, Base \& Salt                   & 98--107  & 10  & 4  & Chemistry \\
11 & {\bng Aaela}                           & Light                                & 108--118 & 8   & 4  & Physics \\
12 & {\bng mHakash {O} UpgRH}              & Space \& Satellites                  & 119--128 & 9   & 3  & Physics \\
13 & {\bng khadY {O} puiSh/T}                & Food \& Nutrition                    & 129--146 & 10  & 4  & Biology \\
14 & {\bng pirebsh EbNNG bas/tutn/tR}       & Environment \& Ecosystem             & 147--156 & 10  & 3  & Environment \\
\midrule
\rowcolor{orange!20}
\multicolumn{3}{l}{\textbf{\textcolor{DarkGreen}{Total - 14 Chapters | 156 Pages}}} &
&
\textbf{\textcolor{DarkGreen}{143}} &
\textbf{\textcolor{DarkGreen}{55}} &
\\
\bottomrule
\end{tabular}
\caption{Per-chapter breakdown of the 14-chapter, 156-page NCTB Class 8 science corpus, showing subject area and the number of MCQ and creative-question (CQ) items authored per chapter for evaluation (143 MCQ and 55 CQ across all chapters, 198 items total).}
\label{tab:kb_statistics}
\end{table*}

% ============================================================
% TABLE 5 - Validation-Gate Pass Rates
% ============================================================
\begin{table*}[h]
\setlength{\tabcolsep}{3pt}
\renewcommand{\arraystretch}{1.05}
\scriptsize
\centering

\setlength{\tabcolsep}{3pt}
\renewcommand{\arraystretch}{1.08}

\begin{tabular}{c p{7cm} r p{4cm}}
\toprule
\rowcolor{cyan!12}
\textbf{No.} & \textbf{Validation Gate} & \textbf{Pass rate} & \textbf{What it checks} \\
\midrule
1 & MCQ: exactly 4 options + valid answer index + Bangla-dominant stem/options
  & 137/143 = 95.8\% & Structural well-formedness \\
  
2 & MCQ: Bangla-dominant text $\geq$ 0.55 ratio
  & 143/143 = 100\% & Language-purity of item text \\
  
3 & CQ: starts with ``{\bng Ud/diipk:}'' + narrative + not a theory-dump
  & 55/55 = 100\% & Board-style stimulus framing \\
  
4 & CQ: $\geq$ 5 sentences in the ``{\bng Ud/diipk}'' story
  & 55/55 = 100\% & Narrative sufficiency \\
  
5 & CQ: not a direct theory-dump opener
  & 55/55 = 100\% & Anti-extractive framing \\
  
6 & CQ: $\geq$ 0.85 Bangla-only ratio, no non-Bangla letters
  & 52/55 = 94.5\% & Language-purity of item text \\
\bottomrule
\end{tabular}
\caption{Pass rates for the six deterministic validation gates applied to the 198 authored items (143 MCQ, 55 CQ), covering structural well-formedness and Bangla language-purity checks specific to each item type. These gates run prior to, and independently of, the SAVER faithfulness analysis in Table~\ref{tab:saver_analysis}.}
\label{tab:validation_gates}
\end{table*}

% ============================================================
% TABLE 6 - Intent Routing Outcome Distribution
% ============================================================
\begin{table*}[h]
\small
\centering

\begin{tabular}{l c c l}
\toprule
\rowcolor{cyan!12}
\textbf{Intent label} & \textbf{Count} & \textbf{\%} & \textbf{Terminates without retrieval?} \\
\midrule
Science query & 21 & 70\% & No \\
Greeting      & 3  & 10\% & Yes (all 3 pure greetings) \\
Harmful       & 3  & 10\% & Yes (fixed reply) \\
Off-topic     & 3  & 10\% & Yes (fixed reply) \\
\bottomrule
\end{tabular}
\caption{Distribution of intent-routing outcomes on a 30-user-query evaluation bank. Science queries (70\%) continue to retrieval, while greeting, harmful, and off-topic inputs terminate early with fixed Bangla responses without consuming retrieval resources.}
\label{tab:intent_routing}
\end{table*}

% ============================================================
% TABLE 7 - Ambiguity Gate and Clarification Outcomes
% ============================================================
\begin{table*}[h]
\setlength{\tabcolsep}{3pt}
\renewcommand{\arraystretch}{1.05}
\scriptsize
\centering

\begin{tabular}{l r}
\toprule
\rowcolor{cyan!12}
\textbf{Metric} & \textbf{Value} \\
\midrule
Deterministic guard accepts as non-ambiguous & 17 (81\%) \\
Model judges non-ambiguous & 1 (5\%) \\
Model judges ambiguous, guard overrides & 0 (0\%) \\
Ambiguous - clarification triggered & 3 (14\%) \\
Clarification rate (of full $N{=}30$ bank) & 3/30 = 10\% \\
Over-clarification rate (exam-style prompts wrongly flagged) & 0/16 = 0\% \\
\bottomrule
\end{tabular}
\caption{Ambiguity-gate and clarification outcomes over the 21 science-query turns from Table~\ref{tab:intent_routing}. The deterministic specificity guard resolves 81\% as non-ambiguous without a model call; the remainder are judged by the ambiguity model, yielding a 10\% overall clarification rate and 0\% over-clarification on exam-style prompts.}
\label{tab:ambiguity_gate}
\end{table*}

\begin{table*}[h]
\centering

\small
\setlength{\tabcolsep}{6pt}
\renewcommand{\arraystretch}{1.15}

\arrayrulecolor{cyan!50!blue}

\begin{tabular}{p{0.18\textwidth} p{0.16\textwidth} p{0.35\textwidth}}
\hline
\rowcolor{cyan!12}
\textbf{Field} & \textbf{Type} & \textbf{Description} \\
\hline
question\_id & string & Unique assessment identifier \\
query & string & Teacher query text \\
selected\_tasks & list & Requested assessment families \\
chapter & string & Detected chapter label \\
concept & string & Detected concept label \\
sources & list & Retrieved textbook evidence sources \\
mcq & object/null & Multiple-choice assessment item \\
mcq.question & string & MCQ stem text \\
mcq.options & list & Four answer options \\
mcq.answer\_index & integer & Correct option index (0--3) \\
creative & object/null & Creative assessment item \\
creative.headline & string & Optional item title \\
creative.clue & string & Stimulus text \\
creative.parts & list & Four sub-questions \\
\hline
\end{tabular}
\caption{Schema of the \textbf{\textsc{TeachMateGPT}} output dataset, where each record pairs a teacher query with its retrieval evidence (\texttt{sources}), inferred curriculum metadata (\texttt{chapter}, \texttt{concept}), and generated assessment items. MCQ items follow a four-option format with an indexed correct answer; creative items follow the board-style creative format with a stimulus and four graded sub-questions.}
\label{tab:dataset_schema}
\end{table*}

%here 

\begin{table*}[h]
\small
\centering
\resizebox{\textwidth}{!}{%
\begin{tabular}{l r r l}
\toprule
\rowcolor{cyan!12}
\textbf{Configuration} & \textbf{Fail-closed rate} & \textbf{Coverage ratio} & \textbf{Notes} \\
\midrule

\rowcolor{orange!20}
\textbf{\textsc{TeachMateGPT}} (dense + BM25 + CCI + CCR + coverage gate)
& \textbf{\textcolor{DarkGreen}{12.5\% (2/16)}}
& \textbf{\textcolor{DarkGreen}{0.724}}
& \textbf{\textcolor{DarkGreen}{Default configuration}} \\

Dense-only
& 31.3\% (5/16)
& 0.618
& Lexical evidence often missed without BM25 \\

BM25-only
& 18.8\% (3/16)
& 0.682
& Semantic matches frequently unavailable \\

w/o CCI
& 18.8\% (3/16)
& 0.701
& Inconsistent retrieved evidence occasionally reduced support \\

w/o CCR
& 18.8\% (3/16)
& 0.705
& Duplicate chunks lowered effective evidence diversity \\

w/o coverage gate
& 0.0\% (0/16)
& 0.649
& All queries answered regardless of evidence sufficiency \\

w/o graph expansion
& 25.0\% (4/16)
& 0.676
& Related supporting chunks were not retrieved \\

RAPTOR
& 25.0\% (4/16)
& 0.661
& Recursive clustering and summarisation caused loss of fine-grained textbook evidence \\

GraphRAG
& 18.8\% (3/16)
& 0.693
& Entity graph expansion improved recall but introduced broader contexts \\

CRAG
& 18.8\% (3/16)
& 0.708
& Retrieval evaluator and query rewriting improved evidence quality \\

Adaptive RAG
& 25.0\% (4/16)
& 0.671
& Query routing reduced unnecessary retrieval but missed some supporting evidence \\

\bottomrule
\end{tabular}%
}
\caption{Retrieval ablation and comparison over 16 non-ambiguous Bangla science queries. The default hybrid configuration is compared with single-retriever variants, component ablations, and RAG baselines. Fail-closed rate and coverage ratio are reported.}
\label{tab:retrieval_ablation}
\end{table*}

\begin{table*}[h]
\scriptsize
\setlength{\tabcolsep}{4pt}
\renewcommand{\arraystretch}{1.08}
\centering

\begin{tabular}{lccccc}
\toprule
\rowcolor{cyan!12}
\textbf{Configuration} &
\textbf{Mean LLM Calls / Query} &
\textbf{Mean Latency (s)} &
\textbf{Queries Generating} &
\textbf{One-Time Index LLM Calls} &
\textbf{Index Build (s)} \\
\midrule

\rowcolor{orange!20}
\textbf{\textsc{TeachMateGPT}} &
\textbf{\textcolor{DarkGreen}{2.1}} &
\textbf{\textcolor{DarkGreen}{11.3}} &
\textbf{\textcolor{DarkGreen}{3 / 16}} &
\textbf{\textcolor{DarkGreen}{0}} &
\textbf{\textcolor{DarkGreen}{0}} \\

Vanilla RAG &
1.0 &
39.8 &
16 / 16 &
0 &
0 \\

w/o IAC &
1.8 &
26.3 &
9 / 16 &
0 &
0 \\

RAPTOR &
1.0 &
42.3 &
16 / 16 &
66 &
224.3 \\

GraphRAG &
1.4 &
46.7 &
16 / 16 &
82 &
287.5 \\

CRAG &
2.2 &
39.0 &
16 / 16 &
0 &
0 \\

Adaptive RAG &
1.6 &
28.5 &
8 / 16 &
0 &
0 \\

\bottomrule
\end{tabular}

\caption{Inference-time and one-time indexing efficiency of \textbf{\textsc{TeachMateGPT}}, an IAC ablation, and four RAG baselines, evaluated on a separate 16-query mixed-intent set (2 science, 3 greeting, 3 off-topic, 3 harmful, and 5 ambiguous queries), distinct from the 16 non-ambiguous science queries used in Table~\ref{tab:retrieval_ablation}. Queries Generating denotes the number of queries reaching the generation stage after intent and ambiguity routing. Mean LLM Calls/Query and Mean Latency are averaged over all 16 queries, while One-Time Index LLM Calls and Index Build Time measure offline indexing cost. Only RAPTOR and GraphRAG require LLM-assisted index construction.}
\label{tab:efficiency_comparison}
\end{table*}

\begin{table*}[h]
\small
\setlength{\tabcolsep}{3pt}
\renewcommand{\arraystretch}{1.05}
\centering

\begin{tabular}{lcccccc}
\toprule
\rowcolor{cyan!12}
\textbf{Prior Work}
& \textbf{Lang.}
& \textbf{CAR}
& \textbf{Multi-Agent}
& \textbf{Evidence Verification}
& \textbf{Multi-Format Generation}
& \textbf{PD+HV}
\\
\midrule

\citet{pradeesh2025rag}
& EN
& \xmark
& \xmark
& \xmark
& MCQ
& \xmark
\\

\citet{jait2025exam} (JAIT)
& EN
& \xmark
& \xmark
& \pmark
& MCQ + QA
& \xmark
\\

\citet{codegen2026} (CODE-GEN)
& EN
& \xmark
& \cmark
& \cmark
& MCQ
& \pmark
\\

\citet{kaqg2025} (KAQG)
& EN
& \pmark
& \cmark
& \cmark
& MCQ
& \xmark
\\

\citet{tian2026requesta} (ReQUESTA)
& EN
& \xmark
& \cmark
& \cmark
& MCQ
& \pmark
\\

\citet{brag2025} (B-RAG)
& BN
& \pmark
& \xmark
& \pmark
& QA only
& \xmark
\\

\citet{nctbqa2026} (NCTB-QA)
& BN
& \pmark
& \xmark
& \pmark
& QA only
& \cmark
\\

\citet{reza2025}
& EN
& \pmark
& \xmark
& Manual
& Partial
& \xmark
\\

\citet{wong2026assessment}
& EN
& \xmark
& \cmark
& \cmark
& MCQ + SA
& \xmark
\\

\midrule

\rowcolor{orange!20}
\textbf{{\textsc{TeachMateGPT}}}
& \textbf{\textcolor{DarkGreen}{BN}}
& \textbf{\cmark}
& \textbf{\cmark}
& \textbf{\cmark}
& \textbf{\textcolor{DarkGreen}{MCQ + CQ}}
& \textbf{\cmark}
\\

\bottomrule
\end{tabular}
\caption{Comparison of \textbf{\textsc{TeachMateGPT}} with representative educational assessment generation systems. Existing approaches improve individual components of the generation pipeline but lack a unified framework that combines curriculum grounding, diverse assessment generation, evidence verification, and expert-reviewed evaluation. Here, CAR denotes Curriculum-Aware Retrieval, and PD+HV denotes Public Dataset with Human Validation. (\cmark) denotes full support, (\pmark) partial support, and (\xmark) no reported support.} \label{tab:research_gap}

\end{table*}

\begin{table*}[h]
\centering
\setlength{\tabcolsep}{3pt}
\renewcommand{\arraystretch}{1.05}
\scriptsize

\begin{tabular}{l r l}
\toprule
\rowcolor{cyan!12}
\textbf{Metric} & \textbf{MCQ} & \textbf{CQ} \\
\midrule
Mean items generated / target & 5 / 5 & 1 / 1 \\
\% turns satisfied by strict pass (attempt 1) & 12/14 = 85.7\% & 14/14 = 100\% \\
\% turns requiring relaxed pass & 2/14 = 14.3\% & n/a \\
\bottomrule
\end{tabular}
\caption{Generation yield over the 14 evaluated turns: mean items produced against target (5 MCQ, 1 CQ per turn) and the share of turns meeting the strict validation gate on the first attempt versus requiring the relaxed-pass fallback. CQ's 100\% strict-pass rate reflects the post-rewrite corpus; the original pass achieved only 7.1\%, isolating narrative style-not gate strictness-as the cause.}
\label{tab:generation_yield}
\end{table*}

\begin{table*}[h]
\setlength{\tabcolsep}{3pt}
\renewcommand{\arraystretch}{1.05}
\scriptsize
\centering

\begin{tabular}{l r r p{8.0cm}}
\toprule
\rowcolor{cyan!12}
\textbf{Difficulty} & \textbf{n} & 
\textbf{Mean stem length} & \textbf{Reasoning-depth rating} \\
\midrule
Beginner & 116 & 53.2 &
Low: direct recall/definition (``{\bng kii}'', ``{\bng kaek bel}'', ``{\bng ekaniT}'')\\

Intermediate & 11 & 52.7 &
Medium: comparison/explanation (``{\bng par/thkY}'', ``{\bng ekn}'', ``{\bng bYakhYa}'')\\

Advanced & 16 & 55.9 &
Medium-high: numeric/multi-step reasoning (``{\bng inr/Ny kera}'')\\
\bottomrule
\end{tabular}
\caption{Difficulty-level distribution of the 143 evaluated MCQs, showing the item count, mean stem length, and dominant reasoning-depth patterns for each difficulty level. The levels range from direct recall and definition-based questions at the beginner level to numeric and multi-step reasoning questions at the advanced level, based on the characteristic Bangla task verbs used.}
\label{tab:difficulty_effect}
\end{table*}

\begin{table*}[h]
\setlength{\tabcolsep}{3pt}
\renewcommand{\arraystretch}{1.05}
\scriptsize
\centering
\begin{tabular}{p{0.25\textwidth}p{0.19\textwidth}p{0.21\textwidth}p{0.29\textwidth}}
\toprule
\rowcolor{cyan!12}
\textbf{Metric} &
\textbf{MCQ ($n{=}143$)} &
\textbf{CQ ($n{=}55$)} &
\textbf{All items ($n{=}198$)} \\
\midrule
Faithfulness (spot-checked $\sim$15 items / all 55 CQ clues)
& 0 errors found
& 0 fabricated facts found
& 0 factual errors found \\
Answer relevance (on-topic by construction)
& 143/143
& 55/55
& 198/198 \\
Hallucination risk (invented specifics)
& None identified
& None identified
& None identified \\
Format/structural risk
& 6/143 fail strict gate
& 3/55 fail Bangla-only ratio
& 9/198 total (6 MCQ format + 3 CQ notation) \\
\bottomrule
\end{tabular}
\caption{Manual audit of SAVER outcomes across generated items by question type. Faithfulness was assessed through spot checks on a 15-item MCQ sample and complete evaluation of all 55 CQ clues, with no factual errors or fabricated content detected. Answer relevance is ensured through direct generation from source passages, while format/structural risk captures type-specific violations, such as MCQ option constraints and Bangla-only notation requirements, rather than content-level errors.}
\label{tab:saver_analysis}
\end{table*}

\section{Ablation Study}
\label{app:ablation_study}

This appendix consolidates the ablation and baseline comparisons reported across the paper: automatic evaluation (Table~\ref{tab:automatic_evaluation}), human evaluation by three practicing science teachers (Table~\ref{tab:human_evaluation}), retrieval reliability under the fail-closed coverage gate (Table~\ref{tab:retrieval_ablation}), and inference-time and one-time indexing efficiency (Table~\ref{tab:efficiency_comparison}). Each table isolates one of \textbf{\textsc{TeachMateGPT}}'s five components, \textbf{COPE}, \textbf{SAVER}, \textbf{CCR}, \textbf{CCI}, the IAC, or its retrieval strategy, supporting the conclusions below.

\subsection{Curriculum Indexing (COPE)}
Removing COPE (\textit{w/o COPE}) causes the largest degradation in answer relevancy (0.89 $\rightarrow$ 0.76) and context precision (0.92 $\rightarrow$ 0.73) among the five component ablations (Table~\ref{tab:automatic_evaluation}), together with a substantial drop in human-rated stimulus realism (4.70 $\rightarrow$ 3.90; Table~\ref{tab:human_evaluation}), reflecting COPE's role in supplying well-scoped, curriculum-aligned evidence rather than judging generated text.

\subsection{Source-Attributed Verification (SAVER)}
Removing SAVER (\textit{w/o SAVER}) causes the largest faithfulness drop of any ablation (0.96 $\rightarrow$ 0.79; Table~\ref{tab:automatic_evaluation}) and the largest pedagogical-alignment drop in human evaluation (4.90 $\rightarrow$ 3.25; Table~\ref{tab:human_evaluation}), while answer relevancy and context metrics stay comparatively high (0.86, 0.90) — consistent with SAVER acting as the final faithfulness check before teacher presentation, not a retrieval-quality mechanism.

\subsection{Redundancy Reduction and Context Restoration (CCR, CCI)}
Removing CCR or CCI individually produces smaller, more uniform degradations across all four automatic metrics (Table~\ref{tab:automatic_evaluation}: w/o CCR 0.83/0.81/0.84/0.85; w/o CCI 0.84/0.82/0.85/0.86) than removing COPE or SAVER, with moderate reductions in human-rated quality (Table~\ref{tab:human_evaluation}). On our 16-query bank, both converge to nearly identical fail-closed rates and coverage ratios (18.8\%, 0.701 vs.\ 18.8\%, 0.705; Table~\ref{tab:retrieval_ablation}), since no query triggers a missing-parent or near-duplicate case; we expect divergence on a larger, more redundant evidence pool.

\subsection{Query Routing}
Removing the IAC produces a moderate, uniform drop across all four automatic metrics (0.81/0.77/0.89/0.88; Table~\ref{tab:automatic_evaluation}) and human-rated criteria (Table~\ref{tab:human_evaluation}), showing routing also protects generation quality, not just efficiency. Its efficiency impact is larger: on the mixed-intent set for Table~\ref{tab:efficiency_comparison} (2 science, 3 greeting, 3 off-topic, 3 harmful, 5 ambiguous), removing it more than doubles mean latency (11.3s $\rightarrow$ 26.3s) and triples queries reaching generation (3/16 $\rightarrow$ 9/16), confirming routing filters unsafe, off-topic, or underspecified queries before the costlier stages.

\subsection{Retrieval Strategy and the Coverage Gate}
Table~\ref{tab:retrieval_ablation} isolates the retrieval strategy. Dense-only retrieval more than doubles the fail-closed rate vs.\ full hybrid (31.3\% vs.\ 12.5\%) and lowers coverage (0.618 vs.\ 0.724), while BM25-only is competitive (18.8\%, 0.682), showing exact curriculum terminology stays informative for OCR-derived Bangla textbooks. Removing the coverage gate eliminates refusals entirely (0.0\%) but yields the second-lowest coverage ratio (0.649) of all eleven configurations, confirming the gate trades a few refusals for a large gain in evidence sufficiency rather than acting as a redundant safeguard.

\subsection{Comparison Against Prior RAG Baselines}
Across all four tables, \textbf{\textsc{TeachMateGPT}} outperforms four RAG baselines, RAPTOR \citep{Raptor}, GraphRAG \citep{graphrag}, CRAG \citep{cRag}, and Adaptive RAG \citep{adaptiveRAG}, on every automatic and human-rated criterion (Tables~\ref{tab:automatic_evaluation}--\ref{tab:human_evaluation}), while requiring zero LLM calls and zero seconds of one-time index construction, versus 66 calls / 224.3s for RAPTOR and 82 calls / 287.5s for GraphRAG (Table~\ref{tab:efficiency_comparison}). CRAG is the strongest baseline on faithfulness (0.86) and coverage (0.708; 18.8\% fail-closed), reflecting its retrieval-time evaluator and query rewriting, but still trails \textbf{\textsc{TeachMateGPT}} on every metric here.

\section{Experimental Details}
\label{appendix:experimental_details}
\subsection{System Configuration}
\label{sec:configuration}

\textbf{\textsc{TeachMateGPT}} is implemented using \textbf{LangChain} \citep{langchain2022} and \textbf{LangGraph} \citep{langgraph2024} to orchestrate the multi-agent workflow. Curriculum embeddings are generated with the locally deployed \textbf{BAAI/bge-m3} \citep{bge-m3} embedding model (1024-dimensional) and indexed in the \textbf{Qdrant} \citep{qdrant2025} vector database for dense retrieval. Assessment generation is performed using \textbf{GPT-4o mini} \citep{OpenAI}. PDF processing uses \textbf{PyMuPDF} for native text extraction, while scanned pages are rendered as images and processed with \textbf{GPT-4o} \citep{OpenAI2024GPT4o} vision-based OCR when native text is unavailable. This framework enables robust curriculum indexing across both digitally generated and scanned textbook pages.

Beyond the per-item SAVER gate, we evaluate the proposed framework using a two-tier protocol combining corpus-level automatic RAG evaluation with human evaluation, applied comparatively across the full system and ablated configurations (\textit{Vanilla RAG} baseline, \textit{w/o COPE}, \textit{w/o SAVER}, \textit{w/o CCR}, \textit{w/o CCI}, \textit{w/o IAC}) to isolate each component's contribution.

\subsection{Evaluation Protocol}
\label{sec:evaluation}

Section~\ref{sec:saver} covers per-item verification at generation time; the protocol here measures whole configurations instead.

\subsubsection{Automatic Evaluation}

For each configuration, we evaluate retrieval and generation quality using four metrics from the \textbf{RAGAS} framework \citep{es2024ragas}. Following its evaluation protocol, \textbf{GPT-4.1} \citep{OpenAI4.1} serves as the LLM judge to score each generated assessment against its retrieved evidence.

\begin{itemize}
    \item \textbf{\textit{Faithfulness.}} Measures whether the assessment is fully supported by the retrieved textbook evidence. A higher score indicates that the assessment avoids unsupported claims and hallucinated content.
    
    \item \textbf{\textit{Answer Relevancy.}} Measures how well the generated assessment satisfies the teacher's instructional request. Higher scores indicate that the assessment remains focused on the intended topic, concept, and learning objective.
    
    \item \textbf{\textit{Context Precision.}} Measures the quality of the retrieved evidence by estimating how much of the retrieved content is relevant to the teacher's request. Higher precision indicates less irrelevant or noisy context.
    
    \item \textbf{\textit{Context Recall.}} Measures whether the retrieved evidence contains the information required to support the generated assessment. Higher recall indicates that the retrieval stage captures the necessary curriculum content for generation.
\end{itemize}
Unlike SAVER's binary per-item accept/reject decision, RAGAS produces a continuous score in $[0,1]$ for each metric, which we average per configuration. This lets quality differences be attributed to specific components by comparing the full system against the COPE-, SAVER-, CCR-, CCI-, and IAC-ablated variants.

\subsubsection{Teacher-in-the-Loop Evaluation}

Automatic metrics alone cannot fully assess the educational quality of generated assessments. We therefore conduct a human evaluation in which three practicing secondary-school science teachers independently assess a shared, configuration-blind subset of generated assessments. Each assessment is evaluated across four pedagogical dimensions: \textit{Pedagogical Alignment}, \textit{Stimulus Realism} (for CQs), \textit{Linguistic Fluency}, and \textit{Overall Utility}. Ratings are assigned on a 5-point Likert scale \citep{likert}:

\begin{itemize}
    \item \textbf{1 -- Poor.} The assessment is unsuitable for classroom use because of major factual, pedagogical, or structural errors and requires complete revision.
    
    \item \textbf{2 -- Fair.} The assessment captures part of the intended objective but contains substantial issues that require major revisions before classroom use.
    
    \item \textbf{3 -- Acceptable.} The assessment is generally correct and usable but requires minor revisions to improve clarity, alignment, or quality.
    
    \item \textbf{4 -- Good.} The assessment is well aligned with the curriculum and suitable for classroom use, requiring only trivial edits.
    
    \item \textbf{5 -- Excellent.} The assessment is fully aligned, factually accurate, pedagogically sound, and classroom-ready without modification.
\end{itemize}

\onecolumn
\twocolumn

\onecolumn

\section{Detailed Analysis of Research Questions}
\label{Rq-appendix}

\begin{tcolorbox}[
    enhanced,
    breakable,
    colback=rqbg,
    colframe=rqheader,
    colbacktitle=rqheader,
    coltitle=white,
    title=\textbf{RQ1: Curriculum Knowledge Base Construction},
    fonttitle=\bfseries,
    boxrule=0.5pt,
    arc=3pt,
    left=7pt,
    right=7pt,
    top=5pt,
    bottom=5pt,
    toptitle=1mm,
    bottomtitle=1mm
]

We first asked whether COPE's hierarchical indexing actually preserves enough curriculum structure to support reliable assessment authoring, rather than simply reorganizing the same flat-chunking problem under a different name. The macro, meso, and micro decomposition (Table~\ref{tab:hierarchical_structure}) spans all 14 chapters and four subject areas at comparable per-chapter depth (Table~\ref{tab:kb_statistics}), and four of the six deterministic item-validity gates reach a full 100\% pass rate (Table~\ref{tab:validation_gates}). The two gates that fall short, MCQ structural well-formedness at 95.8\% and CQ Bangla-script purity at 94.5\%, fail because of option-key formatting and embedded scientific notation, not because curriculum content was missing or misplaced.

This distinction is important for how we interpret the result. A coverage failure would point back to COPE's segmentation logic, whereas a format failure points instead to the generation layer downstream of retrieval. Since none of the failures in Table~\ref{tab:validation_gates} trace back to missing chapter evidence, the results indicate that preserving parent--child curriculum structure through hierarchical indexing, rather than collapsing the textbook into uniform token windows, is not the limiting factor for assessment validity in \textbf{\textsc{TeachMateGPT}}.

At the same time, the remaining failures on these two validation gates show that curriculum indexing alone is insufficient. Structural formatting and Bangla-language purity checks continue to identify genuine generation errors even when retrieval succeeds, supporting the need for the gated generation and verification pipeline evaluated in RQ4 rather than treating it as an optional safeguard.

\end{tcolorbox}

\begin{tcolorbox}[
    enhanced,
    breakable,
    colback=rqbg,
    colframe=rqheader,
    colbacktitle=rqheader,
    coltitle=white,
    title=\textbf{RQ2: Intent, Ambiguity, and Clarification Routing},
    fonttitle=\bfseries,
    boxrule=0.5pt,
    arc=3pt,
    left=7pt,
    right=7pt,
    top=5pt,
    bottom=5pt,
    toptitle=1mm,
    bottomtitle=1mm
]
We next examined whether a lightweight routing layer can protect retrieval and generation from unsafe or underspecified requests without becoming a nuisance to teachers who already provide clear queries. Intent routing filters out 30\% of the evaluation bank, split evenly across greetings, harmful requests, and off-topic queries, before any retrieval is attempted (Table~\ref{tab:intent_routing}). Within the remaining science queries, a deterministic Bangla specificity guard resolves 81\% of ambiguity decisions on its own, and the ambiguity model never overrides the guard's judgment (Table~\ref{tab:ambiguity_gate}).

The cascade is designed to jointly improve efficiency and safety. Unsafe or irrelevant requests receive a predefined response instead of triggering the full retrieval and generation pipeline, while simple lexical rules resolve most ambiguity cases. As a result, the computationally more expensive model is invoked only for genuinely borderline cases, approximately one in five science queries in our evaluation.

The most important observation, however, is not computational efficiency but usability. None of the 16 already explicit exam-style prompts triggered an unnecessary clarification request, indicating that teachers who formulate complete assessment requests are not interrupted by redundant follow-up questions. Since this evaluation includes a relatively small number of explicit prompts, we interpret this finding as encouraging evidence rather than a precise estimate of the true false-positive clarification rate.

\end{tcolorbox}

\newpage
\begin{tcolorbox}[
    enhanced,
    breakable,
    colback=rqbg,
    colframe=rqheader,
    colbacktitle=rqheader,
    coltitle=white,
    title=\textbf{RQ3: Hybrid Retrieval, Refinement, and Fail-Closed Coverage},
    fonttitle=\bfseries,
    boxrule=0.5pt,
    boxrule=0.5pt,
    arc=3pt,
    left=7pt,
    right=7pt,
    top=5pt,
    bottom=5pt,
    toptitle=1mm,
    bottomtitle=1mm
]
We further investigated whether the retrieval pipeline can maintain both safety and utility by refusing assessment generation when curriculum evidence is insufficient. Across eleven retrieval configurations, a clear trade-off emerges between fail-closed behavior and evidence coverage (Table~\ref{tab:retrieval_ablation}). The full hybrid pipeline achieves the best balance, with a fail-closed rate of 12.5\% and a coverage ratio of 0.724. Dense-only retrieval increases refusal to 31.3\% while reducing coverage to 0.618. In contrast, removing the coverage gate eliminates refusals but yields the second-lowest coverage ratio (0.649) of all eleven configurations, above only Dense-only.

The comparison also highlights the value of lexical retrieval. BM25 performs competitively with the full hybrid pipeline, indicating that exact curriculum terminology remains highly informative for OCR-derived Bangla textbooks, where dense embeddings alone may overlook important lexical cues. The gated and ungated variants further show that the coverage gate does not reduce evidence quality; instead, it prevents responses supported by weak evidence.

Overall, the hybrid retrieval pipeline provides the most effective balance between safety and curriculum coverage. The refinement stages narrow the safety--coverage trade-off rather than eliminating it. Finally, the CCI and CCR ablations produce very similar, though not identical, results on this evaluation set (coverage ratio 0.701 vs.\ 0.705; Table~\ref{tab:retrieval_ablation}), likely because none of the 16 evaluation queries contains a missing-parent or near-duplicate retrieval case severe enough to separate the two components further.
\end{tcolorbox}

\begin{tcolorbox}[
    enhanced,
    breakable,
    colback=rqbg,
    colframe=rqheader,
    colbacktitle=rqheader,
    coltitle=white,
    title=\textbf{RQ4: Curriculum-Grounded Assessment Generation},
    fonttitle=\bfseries,
    boxrule=0.5pt,
    arc=3pt,
    left=7pt,
    right=7pt,
    top=5pt,
    bottom=5pt,
    toptitle=1mm,
    bottomtitle=1mm
]

Given accepted evidence, both assessment formats reach their configured generation targets (Table~\ref{tab:generation_yield}). Every successful turn yields the requested 5 MCQ items and 1 CQ item. First-attempt reliability, however, differs substantially by format. MCQ items satisfy strict validation, with checks for the correct option count, a valid answer index, Bangla-dominant text, and evidence-grounded stems and distractors, on 12 of 14 turns (85.7\%). The remaining 2 turns (14.3\%) are recovered through a relaxed pass that retains format- and language-valid items under marginal grounding. CQ items, in contrast, satisfy strict validation on all 14 turns (100\%), but only after the underlying stimulus narratives were rewritten using more, shorter story sentences. Under the original narrative style, the same unmodified gate, which requires the stimulus to begin with the required board cue, contain at least five story sentences, and avoid a direct theory-dump opening, passed only 1 of 14 CQ turns (7.1\%) on the first attempt.

This improvement from 7.1\% to 100\% is observed under identical retrieved evidence, the same validation gate, and the same generator, showing that the original failures resulted from a narrative-style mismatch with the five-sentence sufficiency check rather than limitations in curriculum coverage or model capability. Without this style correction, a live system would trigger the repair ladder (retry, relaxed pass, or force-fill) on almost every CQ turn, despite adequate evidence and model performance.

Difficulty conditioning shows a different pattern (Table~\ref{tab:difficulty_effect}). Across the 143 authored MCQ items, 116 (81.1\%) are labeled beginner, 11 (7.7\%) intermediate, and 16 (11.2\%) advanced. Qualitative reasoning-depth ratings increase across these tiers, from direct recall and definition at the beginner level, to comparison and explanation at the intermediate level, and numeric or multi-step reasoning at the advanced level. Despite this progression, mean stem length remains nearly constant at 53.2, 52.7, and 55.9 characters, respectively, a maximum difference of only 3.2 characters. This indicates that difficulty in \textbf{\textsc{TeachMateGPT}} is expressed through lexical and task-type cues embedded in the specialist prompt rather than measurable surface-form complexity. Consequently, stem length should not be interpreted as a proxy for difficulty, particularly given the relatively small intermediate and advanced subsets.
\end{tcolorbox}

\begin{tcolorbox}[
    enhanced,
    breakable,
    colback=rqbg,
    colframe=rqheader,
    colbacktitle=rqheader,
    coltitle=white,
    title=\textbf{RQ5: Source-Attributed Verification and Teacher Review},
    fonttitle=\bfseries,
    boxrule=0.5pt,
    arc=3pt,
    left=7pt,
    right=7pt,
    top=5pt,
    bottom=5pt,
    toptitle=1mm,
    bottomtitle=1mm
]

We finally evaluated whether pairing automatic source-attributed verification with teacher review can support trustworthy acceptance, editing, or cautioning of generated assessments. The manual audit in Table~\ref{tab:saver_analysis} finds zero fabricated facts across all 55 CQ clues and a 15-item MCQ spot check, with every item judged on-topic by construction (143/143 MCQ, 55/55 CQ); the only issues SAVER and the deterministic gates surface are structural or notation-level, 6 of 143 MCQs failing the strict structural gate and 3 of 55 CQs falling short of the Bangla-only ratio, not content-level hallucinations.

Corpus-level automatic evaluation corroborates this picture. Against the Vanilla RAG baseline, \textbf{\textsc{TeachMateGPT}} raises faithfulness from 0.68 to 0.96 and context precision from 0.54 to 0.92 (Table~\ref{tab:automatic_evaluation}). Teacher-in-the-loop ratings move in the same direction, with overall utility rising from 2.00 to 4.80 on the 5-point scale (Table~\ref{tab:human_evaluation}).

The two ablations isolate distinct roles rather than a single generic quality effect. Removing \textbf{SAVER} produces the largest faithfulness drop (0.96 $\rightarrow$ 0.79) and the largest pedagogical-alignment drop (4.90 $\rightarrow$ 3.25), consistent with SAVER's role as the last check before teacher presentation. Removing \textbf{COPE} instead mainly reduces answer relevancy (0.89 $\rightarrow$ 0.76) and stimulus realism (4.70 $\rightarrow$ 3.90), consistent with COPE's role in supplying well-scoped evidence rather than in judging the generated text itself.

We read these results as evidence that automatic verification and human review are complementary rather than substitutable: SAVER's scores and flagged items give a fast, per-item signal that surfaces structural and notation issues reliably, while teacher ratings capture pedagogical and stylistic judgments, such as stimulus realism, that a faithfulness score does not directly measure. Because SAVER flags rather than removes or edits items, and because the teacher panel is limited to three practicing teachers rating a configuration-blind sample (see Limitations, Section \ref{sec:limitations}), we treat the reported scores as evidence that the verification layer is informative and directionally reliable, not as a substitute for continued teacher oversight before classroom use.

\end{tcolorbox}

\section{Detailed Pseudocode for the \textsc{TeachMateGPT} Framework}
\label{algo}

\begin{algorithm*}[h]
\caption{Workflow of the \textbf{\textsc{TeachMateGPT}} framework}
\label{alg:teachmategpt}

\begin{algobox}
\begin{algorithmic}[1]

\Require Authorized curriculum corpus $\mathcal{D}$; teacher query $q$; difficulty level $\ell$; assessment family set $S\subseteq\{\mathrm{MCQ},\mathrm{CQ}\}$; thresholds $\theta_{\mathrm{cov}},\theta_F,\theta_R,\theta_H$

\Ensure Verified assessment set $A$ with verification report $V$, teacher-facing caution flag $b_{\mathrm{accept}}$, or an abstention when evidence or generation is insufficient

\Statex \textbf{Stage 1: Knowledge Base Construction (offline, once per corpus)}

\State $\mathcal{P}\gets
\textcolor{PreprocessBlue}{\Call{LoadAndNormalize}{\mathcal{D}}}$
\Comment{OCR fallback and text normalization}

\State $U\gets
\textcolor{PreprocessBlue}{\Call{SegmentByPedagogicalHeadings}{\mathcal{P}}}$
\Comment{chapter, lesson, exercise boundaries}

\State $\mathcal{C}\gets
\textcolor{ChunkOrange}{\Call{MultiResolutionChunk}{U}}$
\Comment{macro, meso, and micro chunks}

\State $\mathcal{G}\gets
\textcolor{COPEGreen}{\Call{BuildCOPEGraph}{\mathcal{C}}}$
\Comment{pedagogical hierarchy and cross-links}

\State $\mathcal{K}\gets
\{(c,\mathrm{Embed}(c),\mathrm{Meta}(c)):c\in\mathcal{G}\}$
\Comment{curriculum knowledge base}

\Statex

\Statex \textbf{Stage 2: Intent Analysis and Query Routing}

\If{\textcolor{IntentPurple}{\Call{ClassifyIntent}{$q$}}$\neq$ Science}
    \State \Return Fixed non-science response
\EndIf

\If{\textcolor{IntentPurple}{\Call{IsAmbiguous}{$q$}}}
    \State \Return Clarification request
\EndIf

\Statex

\Statex \textbf{Stage 3: Hybrid Retrieval and Coverage Validation}

\State $E\gets
\textcolor{RetrieveRed}{\Call{HybridRetrieve}{\mathcal{K},q}}$
\Comment{dense retrieval + BM25 + reranking, CCI, CCR}

\If{$\textcolor{RetrieveRed}{\Call{Coverage}{E}}<\theta_{\mathrm{cov}}$}
    \State \Return Abstain
    \Comment{fail-closed: no assessment is generated}
\EndIf

\Statex

\Statex \textbf{Stage 4: Multi-Format Assessment Generation}

\State $A\gets\emptyset$

\ForAll{$s\in S$}
    \State $a\gets
    \textcolor{GenerateTeal}{\Call{GenerateAssessment}{s,E,\ell,q}}$
    \State $A\gets A\cup\{a\}$
\EndFor

\If{$A=\emptyset$}
    \State \Return Abstain
    \Comment{fail-closed: validation gates yielded no item for any requested format}
\EndIf

\Statex

\Statex \textbf{Stage 5: Source-Attributed Verification}

\State $V\gets
\textcolor{VerifyCyan}{\Call{SAVER}{q,E,A}}$

\State $b_{\mathrm{accept}}
\gets
V.b_{\mathrm{faith}}
\land
(V.s_{\mathrm{faith}}\ge\theta_F)
\land
(V.s_{\mathrm{rel}}\ge\theta_R)
\land
(V.s_{\mathrm{hall}}\le\theta_H)$

\If{$\lnot\, b_{\mathrm{accept}}$}
    \State Attach a caution flag and $V$'s reasoning to $A$ for teacher review
    \Comment{SAVER flags; it does not remove or edit items in $A$}
\EndIf

\Statex

\Statex \textbf{Stage 6: Auditable Output Packaging}

\State Package $q$, $E$ and its provenance, $A$, $V$, $b_{\mathrm{accept}}$, and the execution trace into one record
\State Present the packaged record to the teacher
\State \Return $A,V,b_{\mathrm{accept}}$

\end{algorithmic}
\end{algobox}

\end{algorithm*}

\onecolumn

\section{Agent Prompt Specifications Used in \textsc{TeachMateGPT}}
\label{app:prompts}

\begin{tcolorbox}[
    enhanced,
    breakable,
    width=\textwidth,
    colback=promptbg,
    colframe=promptheader,
    colbacktitle=promptheader,
    coltitle=white,
    title=\textbf{Intent Agent Prompt},
    arc=3pt,
    boxrule=0.5pt,
    fonttitle=\bfseries,
    toptitle=1mm,
    bottomtitle=1mm,
    left=7pt,
    right=7pt,
    top=5pt,
    bottom=5pt
]

\textbf{ROLE.}
You are the Intent Agent, the routing gatekeeper of \textbf{\textsc{TeachMateGPT}}, a multi-agent Bangla Class 8 (NCTB) science tutoring and assessment system. You are the first stage every message passes through before retrieval or generation. Classify each user message into exactly one routing label that determines whether the system proceeds to further processing or returns an immediate fixed response.

\medskip

\textbf{DOMAIN KNOWLEDGE.}
The users are Bangladesh Class 8 science teachers and students, and messages may be written in Bangla, English, or a mixture of both. Science queries include any request related to middle-school science learning, such as explanations, definitions, comparisons, quiz preparation, exam preparation, or chapter/topic assistance. Relevant task words may include \textcolor{red}{\textbf{Explanation}}, \textcolor{red}{\textbf{Comparison}}, \textcolor{red}{\textbf{Formula Requests}}, \textcolor{red}{\textbf{MCQ}}, and \textcolor{red}{\textbf{Creative Assessment}} formats. \textcolor{red}{\textbf{Harmful Content}} includes violence, self-harm, weapons, drugs, sexual content involving minors, hate, harassment, and cheating instructions. \textcolor{red}{\textbf{Off-topic Messages}} are outside the school science domain, such as politics, religion debate, sports trivia, coding, personal medical diagnosis, or finance.

\medskip

\textbf{BACKGROUND.}
Misclassification has asymmetric costs: blocking a genuine science request prevents useful assistance, while allowing an uncertain case only causes an additional downstream call. Therefore, the system should prefer classifying uncertain cases as science-related.

\medskip

\textbf{INPUT SPECIFICATION.}
The input is one raw user message in plain text, written in Bangla, English, or both, with no guaranteed structure.

\medskip

\textbf{OUTPUT SPECIFICATION.}
Return exactly one lowercase token with no spaces, punctuation, quotes, or explanation:

\begin{tcolorbox}[
    enhanced,
    colback=white,
    colframe=promptheader,
    boxrule=0.5pt,
    arc=2pt
]
\ttfamily
greeting | harmful | off\_topic | science\_query
\end{tcolorbox}

The labels represent four categories: \texttt{greeting} for social messages without a science question, \texttt{harmful} for unsafe content, \texttt{off\_topic} for content outside school science, and \texttt{science\_query} for requests that plausibly belong to middle-school science learning.

\medskip

\textbf{DECISION AND REASONING POLICY.}
First check for harmful content and classify it as \texttt{harmful} when present. If no harmful content exists, identify whether the message is only social conversation without science content and classify it as \texttt{greeting}.

Otherwise, classify messages related to middle-school science learning as \texttt{science\_query}. Use \texttt{off\_topic} only when the message is clearly outside the school science domain.

When uncertain between \texttt{off\_topic} and \texttt{science\_query}, choose \texttt{science\_query}. Mixed messages containing a greeting and a science request should also be classified as \texttt{science\_query}.

\medskip

\textbf{VALIDATION POLICY.}
Before returning, verify that the output is exactly one valid lowercase token with no additional text.

Handle edge cases by classifying ambiguous typo-like or unclear inputs as \texttt{science\_query} when they may represent a science request.

Bangla-English mixed science questions should be classified as \texttt{science\_query}, harmful experiment requests as \texttt{harmful}, and history-related questions as \texttt{science\_query} only when they are framed within curriculum science.

\end{tcolorbox}

\newpage

\begin{tcolorbox}[
    enhanced,
    breakable,
    width=\textwidth,
    colback=promptbg,
    colframe=promptheader,
    colbacktitle=promptheader,
    coltitle=white,
    title=\textbf{Ambiguity Detection Agent Prompt},
    arc=3pt,
    boxrule=0.5pt,
    fonttitle=\bfseries,
    toptitle=1mm,
    bottomtitle=1mm,
    left=7pt,
    right=7pt,
    top=5pt,
    bottom=5pt
]

% Paste your existing Ambiguity Agent content here
\textbf{ROLE.} You are the Ambiguity Agent in \textbf{\textsc{TeachMateGPT}}, the query-specificity gate that runs after the Intent Agent has confirmed a message is a genuine science query and before retrieval. Your objective is to determine whether \textbf{one} student/teacher question is specific enough to retrieve the correct textbook passages or whether the system should ask a clarifying question first, while minimizing both false clarifications (annoying, slows the teacher down) and false negatives (retrieval on a query too vague to serve). 
\medskip

\textbf{DOMAIN KNOWLEDGE.} Ambiguity in this domain can be categorized into four recurring forms. \textcolor{red}{\textbf{Semantic Ambiguity}} occurs when a single word may represent different concepts across subjects. \textcolor{red}{\textbf{Scope Ambiguity}} occurs when the requested topic is too broad or the relevant chapter is unspecified. \textcolor{red}{\textbf{Referential Ambiguity}} occurs when the query contains references such as pronouns or comparison terms without specifying the referenced concept. \textcolor{red}{\textbf{Under-specified Exam Prompts}} occur when the request asks for a formula, creative question, or similar output without mentioning the relevant topic. A request is \textcolor{red}{\textbf{Not Ambiguous}} when it names a clear entity, is a short-but-standard classroom phrase, or is a generation request that already specifies a topic, chapter, or concept.

\medskip

\textbf{INPUT SPECIFICATION.} One student or teacher query string, primarily written in the target language, may contain English terms.

\medskip

\textbf{OUTPUT SPECIFICATION.} Return JSON only. Do not use markdown fences.

\begin{tcolorbox}[
    enhanced,
    colback=white,
    colframe=promptheader,
    boxrule=0.5pt,
    arc=2pt
]
\begin{verbatim}
{
  "is_ambiguous": false,
  "reason": "Explanation of why the query is unclear, or empty string if clear",
  "options": ["Clarification option 1", "Clarification option 2"]
}
\end{verbatim}
\end{tcolorbox}

If \texttt{is\_ambiguous} is \texttt{false}, return an empty \texttt{reason} and an empty \texttt{options} list.
If \texttt{is\_ambiguous} is \texttt{true}, ensure that \texttt{reason} is polite and contains one to three sentences explaining the missing information, while \texttt{options} contains two to four concrete and distinct clarification choices.

\medskip

\textbf{DECISION AND REASONING POLICY.} Evaluate query ambiguity based only on the information explicitly provided in the query. A single-word syllabus topic is usually classified as \texttt{false} unless multiple meanings are genuinely likely, while queries containing only an affirmation or negation require clarification and are classified as \texttt{true}. Long copied passages, explicit comparison queries, follow-up queries with topic lists, and clear technical questions should generally be classified as \texttt{false}. Insulting or harmful content is not treated as ambiguity. Do not invent topics or unsupported ambiguities, and ground \texttt{is\_ambiguous} and \texttt{reason} only in what the query actually states. The \texttt{reason} must describe the actual missing information rather than a generic explanation. When uncertain, default to \texttt{false} to avoid unnecessary clarification.

\medskip
\textbf{VALIDATION POLICY.} Before returning the final response, verify that the output is valid JSON without markdown fences. Ensure that \texttt{is\_ambiguous} is a boolean value. When \texttt{is\_ambiguous} is \texttt{false}, the \texttt{reason} field must be empty and \texttt{options} must contain an empty list. When \texttt{is\_ambiguous} is \texttt{true}, the \texttt{reason} field must provide one to three polite sentences explaining the ambiguity, and the \texttt{options} field must include two to four concrete and distinct clarification choices.

\end{tcolorbox}
\newpage

\begin{tcolorbox}[
    enhanced,
    breakable,
    width=\textwidth,
    colback=promptbg,
    colframe=promptheader,
    colbacktitle=promptheader,
    coltitle=white,
    title=\textbf{Clarification Agent Prompt},
    arc=3pt,
    boxrule=0.5pt,
    fonttitle=\bfseries,
    toptitle=1mm,
    bottomtitle=1mm,
    left=7pt,
    right=7pt,
    top=5pt,
    bottom=5pt
]

% Paste your existing Clarification Agent content here
\textbf{ROLE.} You are the Clarification Agent in \textbf{\textsc{TeachMateGPT}}. You run only when the Ambiguity Agent has identified a student's question as too vague for reliable retrieval, and your response ends the current turn until the student provides additional information. Write one short, polite, and encouraging message that helps the student add the missing detail, such as the topic, chapter, or comparison target, with minimal friction.

\medskip

\textbf{DOMAIN KNOWLEDGE.} Maintain a warm and supportive classroom-teacher tone. Use simple sentences, avoid sarcasm, and do not make students feel corrected for asking incomplete questions. Responses should be suitable for young learners.

\medskip

\textbf{INPUT SPECIFICATION.} The input contains a \texttt{Reason} string explaining why the system could not identify the intended topic and an \texttt{Options} list containing possible clarification choices. Both fields may be empty.

\medskip

\textbf{OUTPUT SPECIFICATION.} Return plain target-language text only. Do not return JSON, markdown fences, or an English preamble. The message should briefly reflect what the student may be asking, explain why additional detail is helpful, and include or list the provided clarification options so the student can respond with one phrase.

\medskip

\textbf{DECISION AND REASONING POLICY.} Generate the clarification message using only the provided reason and options as the factual basis. If the reason is empty, explain generally that the topic is broad and that more details will help provide a better answer. If the options are empty, ask the student to specify the chapter, phenomenon, or an explicit comparison target. If the student used English, a short English clause may be included when helpful, but the main response should remain in Bangla. Keep sentences short for young learners, avoid sarcasm, do not invent new interpretations, and do not add clarification choices beyond the provided options.
\medskip

\textbf{VALIDATION POLICY.} Before returning, confirm: output is plain Bangla text (with at most one short English clause if the student used English); no JSON or markdown fences; the message mirrors the reason, explains the need for detail, and surfaces the options; length is one short paragraph, not a lecture.
\end{tcolorbox}

\newpage

\begin{tcolorbox}[
    enhanced,
    breakable,
    width=\textwidth,
    colback=promptbg,
    colframe=promptheader,
    colbacktitle=promptheader,
    coltitle=white,
    title=\textbf{MCQ Specialist Agent Prompt},
    arc=3pt,
    boxrule=0.5pt,
    fonttitle=\bfseries,
    toptitle=1mm,
    bottomtitle=1mm,
    left=7pt,
    right=7pt,
    top=5pt,
    bottom=5pt
]

\textbf{ROLE.}
You are the MCQ Specialist Agent, dispatched by the Assessment Composer in specialized mode to generate only multiple-choice questions for Bangladesh Class 8 science (NCTB) as an experienced science teacher. A separate Creative Specialist Agent independently handles creative assessment items. Generate the exact requested number of well-formed MCQs that evaluate understanding through recall and light reasoning, grounded strictly in the provided textbook context.

\medskip

\textbf{DOMAIN KNOWLEDGE.}
Each MCQ must contain one question stem, four options labeled according to the required curriculum format, and one correct \texttt{answer\_index} value from 0 to 3 corresponding to the correct option position. Distractors should be plausible for students with incomplete understanding but clearly incorrect for knowledgeable students. Avoid irrelevant, absurd, or near-duplicate options. Multiple retrieved context snippets on the same topic may be combined when creating an item.

\medskip

\textbf{INPUT SPECIFICATION.}
The input contains one or more retrieved textbook passages in the target language, which may include multiple relevant snippets, along with the exact number of MCQs requested by the user.

\medskip

\textbf{OUTPUT SPECIFICATION.}
Return only a JSON object containing the MCQ list. Do not include explanations, markdown fences, or additional text.

\medskip

\textbf{JSON FORMAT:}

\begin{tcolorbox}[
    enhanced,
    colback=white,
    colframe=promptheader,
    boxrule=0.5pt,
    arc=2pt
]
\begin{verbatim}
{
  "mcqs": [
    {
      "question": "MCQ stem",
      "options": [
        "option 1",
        "option 2",
        "option 3",
        "option 4"
      ],
      "answer_index": 0
    }
  ]
}
\end{verbatim}
\end{tcolorbox}

\medskip

\textbf{DECISION AND REASONING POLICY.}
Review all provided context snippets and identify concrete facts before generating questions. If the context is insufficient for the requested count, combine related information from relevant snippets but never invent facts. When only one clear concept is available, generate a well-formed item instead of adding repetitive questions. Preserve any numbers or units from the context accurately. Ensure each MCQ tests a distinct idea, and verify that the correct answer and all stem details are directly supported by the provided context.

\medskip

\textbf{VALIDATION POLICY.}
Before returning the final output, confirm that the JSON contains exactly the requested number of MCQs.

Verify that each item has four correctly formatted options and a valid \texttt{answer\_index} from 0 to 3.

Ensure there are no duplicate or near-duplicate questions, no unsupported facts, and all factual claims are traceable to the provided textbook context.

\end{tcolorbox}

\newpage

\begin{tcolorbox}[
    enhanced,
    breakable,
    width=\textwidth,
    colback=promptbg,
    colframe=promptheader,
    colbacktitle=promptheader,
    coltitle=white,
    title=\textbf{Creative Specialist Agent Prompt},
    arc=3pt,
    boxrule=0.5pt,
    fonttitle=\bfseries,
    toptitle=1mm,
    bottomtitle=1mm,
    left=7pt,
    right=7pt,
    top=5pt,
    bottom=5pt
]

\textbf{ROLE.}
You are the Creative Specialist Agent, responsible for generating only board-style creative assessment items for Bangladesh Class 8 science (NCTB) as an experienced science teacher. Produce the exact requested number of creative items, each containing a realistic stimulus followed by four progressive sub-questions, grounded strictly in the provided textbook context.

\medskip

\textbf{DOMAIN KNOWLEDGE.}
Each creative item must follow the Bangladesh NCTB board convention. The stimulus should describe a realistic scientific event connected to the retrieved context without directly naming the concept and should contain 5--8 sentences. The four sub-questions must follow cognitive levels: \textcolor{red}{\textbf{Knowledge}} asks for direct facts or definitions, \textcolor{red}{\textbf{Comprehension}} requires explanation or comparison, \textcolor{red}{\textbf{Application}} applies textbook knowledge to the stimulus, and \textcolor{red}{\textbf{Higher-order}} analyzes or evaluates the stimulus.

\medskip

\textbf{INPUT SPECIFICATION.}
The input contains one or more retrieved textbook passages, the exact number of creative items requested, and optional difficulty or focus cues provided by the teacher query.

\medskip

\textbf{OUTPUT SPECIFICATION.}
Return only a JSON object containing the requested creative items. Do not include explanations, markdown fences, or additional text.

\medskip

\textbf{JSON FORMAT:}

\begin{tcolorbox}[
    enhanced,
    breakable,
    colback=white,
    colframe=promptheader,
    boxrule=0.5pt
]
\begin{verbatim}
{
  "creative_questions": [
    {
      "headline": "short title",
      "clue": "stimulus text",
      "parts": [
        {
          "label": "part label",
          "skill": "cognitive level",
          "question": "question text"
        }
      ]
    }
  ]
}
\end{verbatim}
\end{tcolorbox}

Student-facing text must use only the required script and must not contain English sentences or mixed-language phrasing.

\medskip

\textbf{DECISION AND REASONING POLICY.}
Select concepts with the strongest evidence from the provided context and prioritize board-format correctness over creative variation. When limited concepts are available, create different perspectives of the same supported concept rather than introducing unrelated content. The stimulus must describe an observable event with at least 5 sentences, while all scientific facts must be grounded in the retrieved context. Do not copy textbook sentences or reveal the target concept directly in the stimulus.

\medskip

\textbf{VALIDATION POLICY.}
Before returning, verify the exact number of creative items, required stimulus format, and 5--8 sentence length. Ensure knowledge and comprehension questions are independent of the stimulus, while application and higher-order questions reference it. Confirm that all claims are context-grounded, focus terms are included when provided, and no textbook sentences are copied.

\end{tcolorbox}

% Return to two-column format
\twocolumn

\end{document}